\documentclass[]{bytedance_seed}
\microtypesetup{expansion=false}

\usepackage[toc, page, header]{appendix}
\usepackage{xspace}
\usepackage{amsmath, amssymb}

\newcommand{\ourwork}{Lucida\xspace}

\usepackage{minitoc}
\usepackage{xcolor}

\newtcolorbox{actionbox}{
  enhanced,
  colback=black!2, colframe=black!25, boxrule=0.4pt, arc=2pt,
  left=6pt, right=6pt, top=3pt, bottom=1pt,
  before skip=6pt, after skip=6pt,
  fontupper=\small,
}

\title{{\fontsize{15}{18}\selectfont Lucida: Parse, Generate, and Place\\for Composable Real-to-Sim Scene Modeling}}

\affiliation[1]{ByteDance Seed}
\affiliation[2]{Peking University}
\affiliation[3]{Zhejiang University}
\authornote{See \hyperref[sec:contributions]{Contributions} section for a full author list.}

\abstract{
Composable scene modeling aims to recover a real indoor scene as complete, editable object assets arranged as observed, giving robot simulation and embodied AI a simulation-ready replica of the real environment whose objects can be manipulated individually. Existing pipelines decompose the task into three steps---parse the observations into instances, generate an asset for each, and place each asset back---but every step presumes an input that a cluttered capture rarely provides: accurate instance geometry, unoccluded views, and assets that accurately match the observations. We propose \ourwork, which keeps this order but redistributes the requirements, so each step consumes only what a real capture reliably provides and precision is reached at the end of the pipeline rather than demanded at its start. \ourwork parses the video into a scene graph whose nodes carry per-instance multi-view evidence, generates a complete asset for each instance from its evidence, and places assets with GizmoAct, a VLM policy that casts placement as multi-turn GUI interaction, manipulating the object's gizmo in a closed loop and deciding itself when alignment is reached. Across scene-level 3D object detection, object pose estimation, and scene reconstruction, \ourwork improves mAP over Boxer by 69\% on R2S-Scene, raises ADD-SB@0.05 from 57.8\% to 83.4\% on CA-1M, and increases scene F-Score from 0.794 for SAM~3D to 0.924.
}

\date{\today}

\checkdata[Project Page]{\url{https://lucida-r2s.github.io/}}

\hypersetup{
  pdftitle={Lucida: Parse, Generate, and Place for Composable Real-to-Sim Scene Modeling},
  pdfauthor={Minghan Qin, Yuang Wang, Xiuyu Yang, Yushi Long, Yujian Zhang, Ruihuan Wang, Kai Ye, Yangang Zhang, Hang Li},
  pdfkeywords={real-to-sim, composable scene modeling, 3D scene reconstruction, object pose estimation, vision-language models}
}

\begin{document}
\maketitle

\section{Introduction}
\begin{figure}[t]
    \centering
    \includegraphics[width=\linewidth]{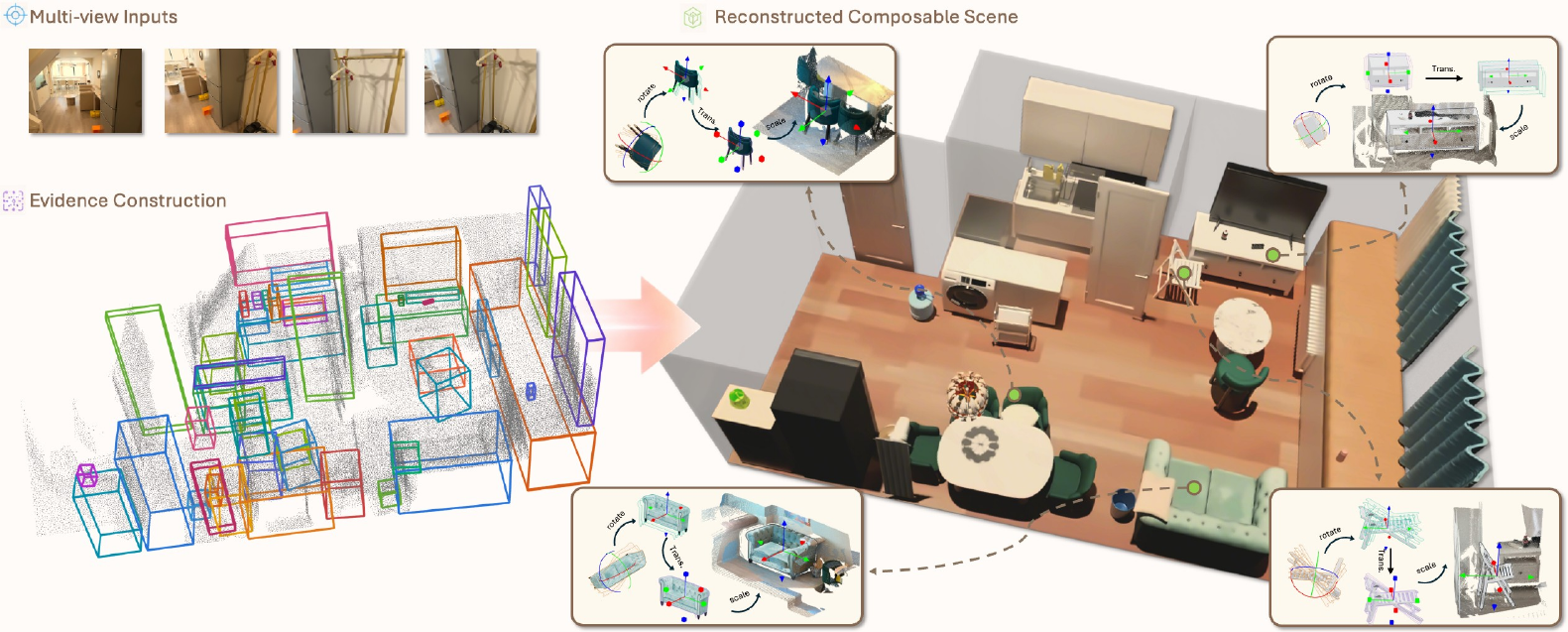}
    \caption{\textbf{Teaser.} Given posed multi-view observations of a real indoor scene, our method constructs object-centric multi-view evidence, generates complete 3D assets, and places them into a composable reconstruction. The insets illustrate closed-loop rotation, translation, and scale refinement of object models against the observed scene geometry.}
    \label{fig:teaser}
\end{figure}

Composable scene modeling recovers a real indoor scene as separable, editable object assets together with their spatial arrangement, the form in which robot simulation, embodied AI, AR/VR, and content creation consume it. Reconstruction by differentiable rendering, as in NeRF \citep{mildenhall2020nerf} and 3D Gaussian Splatting \citep{kerbl20233d}, reproduces the captured room faithfully but as a monolith, and object-compositional extensions \citep{yu2021uorf, kundu2022pnf, siddiqui2022panopticlifting, wu2023objectsdfplusplus} remain tied to the observed geometry, leaving occluded surfaces unrecovered. Real-to-sim systems for robot learning \citep{dai2024acdc, torne2024rialto, chen2024urdformer} and the CAD retrieval-and-alignment methods they build on \citep{avetisyan2018scan2cad, gumeli2021roca} return simulation-ready scenes, but populate the room with library assets, capping fidelity at database coverage. Interactive scene synthesis \citep{yang2023holodeck, yang2024physcene} produces plausible rooms that are grounded in no particular capture. No existing route yields objects that are at once complete, editable assets and faithful to the room that was observed.

Building such a scene decomposes into three steps: \emph{parse} the observations into object instances, \emph{generate} an asset for each, and \emph{place} each asset back into the scene. Strong methods exist for every step, yet each presumes an input that a cluttered capture does not provide. Scene parsing is expected to deliver accurate instance-level estimates such as precise masks, clean per-instance point clouds, or a metric 3D layout, while occlusion leaves object boundaries ambiguous and repeated furniture makes cross-view association and de-duplication unreliable. Generation presumes an unoccluded, centered view of the object \citep{hong2023lrm, tochilkin2024triposr} or a cleanly segmented point cloud of it \citep{siddiqui2026shaper}, neither of which indoor footage readily yields. Placement, typically cast as 6/9-DoF pose estimation \citep{labbe2022megapose, wen2023foundationpose, lee2025any6d}, assumes an asset that agrees geometrically with a segmented observation, while generated assets deviate from the real object and depth is noisy. End-to-end methods do not escape these constraints: predicting an object's model and pose jointly degrades both \citep{sam3dteam2025sam3d, zadaianchuk2026recgen}, and generating all assets of a scene with a single model is limited by scarce paired training data and generalizes poorly \citep{chen2025autopartgen, huang2024midi, meng2025scenegen}. Since the three steps run in a fixed order, an unmet requirement at any step degrades every step after it.

A human modeler working from the same capture satisfies none of these requirements. They review the footage to note what is present, build or retrieve a model for each object, and drop every model into the scene, nudging it against the observation---switching viewpoints, correcting the residual---until the two agree. Precision is reached at the end of this process, not demanded at its start.
We keep the parse--generate--place order but redistribute the requirements along it, so each step consumes only what a real capture reliably provides. Parsing no longer requires a precise instance reconstruction as its output. Instead, it discovers object instances, consolidates their multi-view evidence, and obtains a representative 3D estimate for initializing subsequent generation and grounding. Generation conditions on this per-object evidence rather than on a clean crop or a segmented point cloud, completing the structure and appearance of the target instance under occlusion, which image generation and editing with unified multimodal models \citep{seed2026seedream5pro, openai2026gptimage2, google2025nanobananapro} has made practical. Placement accepts an asset that need not match the observation exactly and, from a coarse initial state and a referring cue, recovers the instance's 9-DoF pose in the point cloud.

We name the resulting system \emph{\ourwork}, after the camera lucida, the drawing aid that superimposes the real scene onto the artist's page. Geometry-aware keyframe selection and object discovery provide candidate object instances and initial 3D estimates. Object-centric full-sequence evidence consolidation then extends and validates each candidate against the full sequence, after which relation-aware scene refinement organizes the consolidated instances into a scene graph and improves their consistency. Its nodes represent object instances and its edges encode spatial relations; each node carries an \emph{evidence bundle} combining selected multi-view observations, partial point-cloud observations, a representative 3D box, and a referring description. The bundle serves both later steps: generation uses it as context to synthesize an occluder-free image of the instance and lift that image into a complete 3D asset, and placement takes from it the coarse initial state for refinement.

We cast the place step as multi-turn GUI interaction: the object is presented as if selected in a 3D editor---the scene point cloud, the asset overlay, its 3D box, and a gizmo exposing the object's local frame---and aligning the asset means issuing a sequence of edits through this interface. \emph{GizmoAct}, a VLM policy, operates the interface in a closed loop. At each turn it reads the rendered state and emits one executable edit, an incremental pose update in the gizmo frame whose translation and scale are expressed in units of the current object size, so the policy never regresses an absolute pose or estimates metric scale. The loop resembles render-and-compare refinement \citep{li2018deepim, lin2020inerf, labbe2022megapose}, but where render-and-compare infers each update by comparing a rendering of the current estimate with the observation and needs an external criterion to decide convergence, GizmoAct predicts both the next edit and when to stop from the rendering alone. We obtain the policy by supervised finetuning of a pretrained VLM on synthetic expert trajectories, followed by reinforcement learning in the same environment used at inference; because training uses assets whose geometry disagrees with the image evidence, initialized at deliberately hard random poses, the policy stays accurate under mismatched assets and coarse initialization.

We evaluate scene parsing and object grounding separately, followed by the complete system they support. For scene-level 3D object detection, the scene graph of the parse step is compared against the prompt-based 3D detectors Boxer \citep{boxer2026} and WildDet3D \citep{wilddet3d2026} on CA-1M \citep{lazarow2024cubify} and on R2S, our real-world indoor benchmark for real-to-sim reconstruction. \ourwork achieves the highest mean mAP in all four evaluation protocols, improving over Boxer from 0.351 to 0.592 on R2S-Scene under the all-annotation protocol and leading on both CA-1M protocols. For object pose estimation, GizmoAct refines coarse layouts on R2S-Object, CA-1M, and ADT \citep{pan2023aria}, raising ADD-SB@0.05 over the strongest baseline from 57.8\% to 83.4\% on CA-1M and from 79.2\% to 92.0\% on R2S-Object, and improving 3D IoU from 0.434 to 0.607 and from 0.500 to 0.719, respectively; a single policy handles initializers with different error profiles without retraining. At the system level, scenes composed by \ourwork achieve a scene F-Score of 0.924, compared with 0.794 for SAM~3D \citep{sam3dteam2025sam3d} and 0.351 for SceneGen \citep{meng2025scenegen}, while improving the object-level F-Score from 0.704 for SAM~3D to 0.736.

Our contributions are summarized as follows:
\begin{itemize}
    \item We recast composable scene modeling so that no step depends on an input a real cluttered capture cannot provide, deferring precision to a closed-loop final step instead of demanding it from the first.
    \item We propose a scene-level 3D object detection method that builds a scene graph with per-instance multi-view evidence, improving mAP over Boxer from 0.351 to 0.592 on R2S-Scene under the all-annotation protocol.
    \item We propose GizmoAct, a VLM policy that casts asset placement as multi-turn GUI interaction, aligns the asset by closed-loop gizmo manipulation, and stops autonomously; trained by supervised finetuning and reinforcement learning, it raises strict-alignment success (ADD-SB@0.05) by up to 25.6 percentage points over the strongest baseline while accepting mismatched assets and coarse initial poses.
    \item We integrate these modules into \ourwork and evaluate it on R2S, our real-world real-to-sim benchmark, at three levels---3D object detection, object pose estimation, and scene reconstruction---where the composed scenes achieve a scene F-Score of 0.924, compared with 0.794 for the strongest baseline.
\end{itemize}

\section{Method}

\begin{figure}[t]
    \centering
    \includegraphics[width=\linewidth]{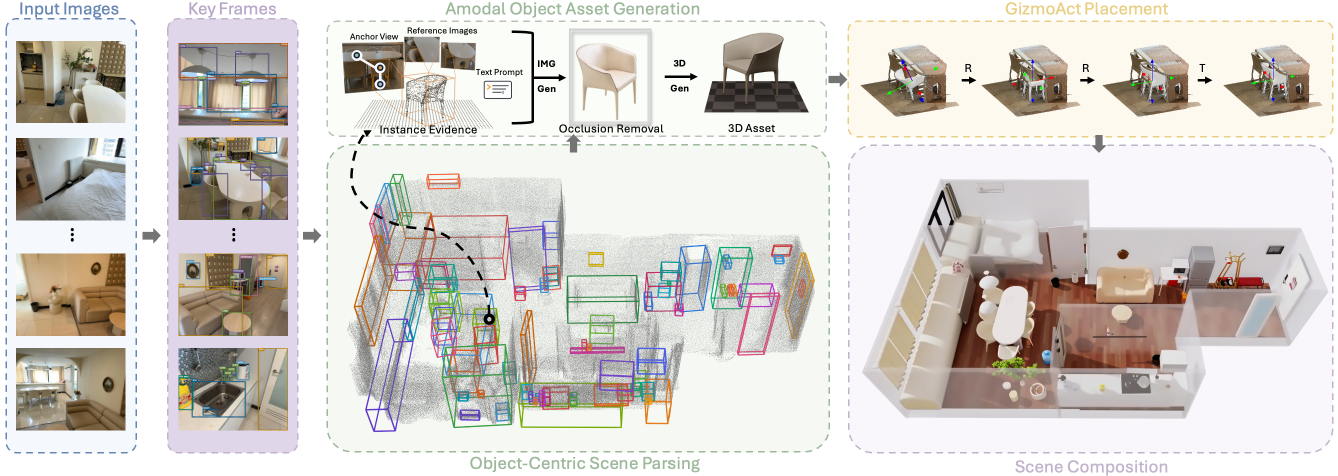}
    \caption{\textbf{Overview of our method.} Given posed scene observations, the parsing
stage constructs an object-centric scene graph with per-object multi-view
evidence and coarse 3D estimates. For each object, the multi-view evidence is
used to synthesize a complete object-centric image, which is then lifted into a
3D asset.  GizmoAct places the asset through closed-loop 9-DoF gizmo manipulation, iteratively refining rotation, translation, and anisotropic scale until it decides that alignment is reached. The placed assets form an editable reconstruction of the captured scene.}
    \label{fig:pipeline}
\end{figure}

\ourwork converts posed RGB(-D) observations of an indoor scene into an
editable, object-level replica: each real instance becomes a complete 3D
asset placed by a 9-DoF pose and organized in a scene graph. As shown in~\Cref{fig:pipeline},
the pipeline follows the parse--generate--place order.
The parse stage
(Sec.~\ref{sec:evidence}) consolidates the multi-view observations into a
scene graph whose nodes carry per-instance \emph{evidence bundles}
$\mathcal{E}_o$, which provide per-object context for the later stages.
The generate
stage (Sec.~\ref{sec:amodal-generation}) synthesizes an occlusion-free
 image from $\mathcal{E}_o$ and lifts it into a 3D asset
$A_o$.
GizmoAct (Sec.~\ref{sec:gizmoact}) then places $A_o$ through
closed-loop gizmo manipulation, starting from a coarse initial pose from
$\mathcal{E}_o$ and refining the pose until it decides alignment is
reached;
optional postprocessing further enforces scene-level constraints
(Sec.~\ref{sec:postprocess}).
The design principle throughout is that each stage consumes only what a
real capture reliably provides, deferring precision to the closed-loop
object placement stage.

\subsection{Multi-View Object-Centric Scene Parsing}
\label{sec:evidence}

The first module parses multi-view scene observations into an object-centric
scene graph that consolidates visual and geometric evidence for both amodal
generation and 3D grounding. Given posed RGB-D observations
$\mathcal{I}=\{(I_i, D_i, K_i, T_i)\}_{i=1}^{N}$, where $I_i$ is an RGB image,
$D_i$ a depth map, $K_i$ the camera intrinsic matrix, and $T_i$ the camera pose,
\ourwork builds a scene graph $G=(V,E)$. For each object $o$, the graph stores
a multi-view evidence bundle
\begin{equation}
    \mathcal{E}_o = \{\mathcal{V}_o, \mathcal{M}_o, \mathcal{P}_o, b_o, c_o\},
\end{equation}
where $\mathcal{V}_o$ contains multi-view observations of the object,
$\mathcal{M}_o$ contains the associated masks or boxes, $\mathcal{P}_o$ contains
partial point-cloud observations, $b_o$ is a representative 3D box, and $c_o$
is a category name or referring description. Edges in $E$ record spatial
relations between objects, such as support, containment, and adjacency.

\paragraph{Geometry-Aware Keyframe Selection and Object Discovery.}
Adjacent frames in a captured sequence are often highly overlapping, making
object discovery on every frame redundant. We first select a sparse set of keyframes from the input sequence. For each
pair of frames $i,j$, we compute a similarity $s(i,j)$ from covisibility and temporal separation. Covisibility measures
geometric redundancy, while the temporal term reduces the influence of distant
frames so that keyframes remain distributed over a long sequence:
\begin{equation}
    \begin{aligned}
    c_{i\to j} &=
    \frac{|\mathcal C_{i\to j}|}{|\mathcal P_i|},\quad
    \mathcal C_{i\to j}
    =
    \left\{
    p\in\mathcal P_i:
    \left|D_{i\to j}(p)-D_j(\pi_j(p))\right|\le\delta
    \right\}, \\
    s(i,j) &=
    \frac{2c_{i\to j}c_{j\to i}}
    {\max(c_{i\to j}+c_{j\to i},\epsilon)}
    \left(
    \lambda+(1-\lambda)
    \exp\left(-\frac{|i-j|}{\tau}\right)
    \right),
    \end{aligned}
\end{equation}
where $\mathcal P_i$ contains sampled 3D points from frame $i$,
$\pi_j(p)$ projects $p$ into frame $j$, $D_{i\to j}(p)$ denotes its
projected depth, $c_{i\to j}$ denotes directional covisibility, $\lambda$ and
$\tau$ control the temporal weighting. We traverse the sequence in
temporal order and retain a frame when its similarity to every selected keyframe falls below a threshold.

On the selected keyframes, a VLM identifies object instances and a 3D
detector~\citep{boxer2026} predicts a box for each object. We group
observations across keyframes based on semantic and geometric consistency
in the common 3D coordinate frame. The grouped observations and their
per-frame 3D boxes form the initial evidence for each candidate object.

\paragraph{Object-Centric Full-Sequence Evidence Consolidation.}
Keyframes are selected according to scene-level covisibility and therefore do
not guarantee the sufficient observations for every object. 
To retrieve additional
observations for each object from the full sequence, we first select a
representative 3D box from its geometrically consistent 3D
boxes across keyframes.
If the 3D boxes from the keyframes are not sufficiently consistent, we propagate the object to nearby frames using video segmentation and
tracking~\citep{ravi2024sam2}, lift the additional observations into 3D, and repeat the selection with the expanded evidence.

We project the representative box into the input frames and use the projected
2D regions as prompts for per-frame 3D box estimation. We validate these observations against the representative box and
observed geometry, retaining the consistent ones as the full-sequence evidence
for the object.

\paragraph{Relation-Aware Scene Refinement.}
Processing objects independently can leave scene-level errors such as
incorrect grouping and spatial inconsistencies. We compare appearance and geometry across objects to correct erroneous merges
and splits. We also infer spatial relations between objects. For a resting object without
valid support in the current graph, we search additional views for the missing
supporting object and add it after multi-view geometric validation. The inferred
relations also guide the correction of spatial inconsistencies between objects
and their surroundings. 

The resulting per-object evidence guides amodal asset generation and provides
object references and coarse spatial initialization for GizmoAct placement.

\subsection{Amodal Object Asset Generation}
\label{sec:amodal-generation}

The second module converts each multi-view evidence bundle into a complete
standalone object asset. In real captures, occlusion and noisy depth often
leave the per-object point-cloud observations incomplete and unreliable for
direct 3D completion. In contrast, multi-view RGB observations reveal
complementary parts of the object across viewpoints. We therefore first use
the multi-view visual evidence to synthesize a complete object-centric image
and then lift the completed object into 3D. Specifically, we select a small
set of reliable and complementary views from $\mathcal{E}_o$, favoring
observations with clear object visibility and geometric agreement with the
representative 3D box while encouraging diversity in viewpoint and occlusion.
To organize the selected views for image editing, we use
Set-of-Mark prompting~\citep{yang2023som} to indicate the target in each view, while a
VLM organizes the selected views into an anchor and references
and produces an editing instruction for amodal completion. The image-editing
model uses these views and the instruction to synthesize a complete, isolated
object image. Finally, an image-to-3D model~\cite{gu2026seed3d20advancinghighfidelity} converts the completed image into
an object asset $A_o$, which is coarsely initialized in the scene using $b_o$
and subsequently refined by GizmoAct in position, orientation, and anisotropic
scale.

\subsection{Agentic 3D Grounding with GizmoAct}
\label{sec:gizmoact}

\subsubsection{Formulation: 3D Grounding as Multi-turn GUI Interaction}

3D grounding is commonly posed as localizing a referred object with a 3D box, yet the box is only one proxy for the spatial state of that object: a 3D model instead yields object pose estimation, a coordinate frame yields orientation estimation, and a planar primitive yields layout estimation. What these tasks share is the operation of aligning a 3D element with the available image and geometry evidence, much as a modeler places and adjusts an element against a reference in a 3D modeling tool. We therefore reformulate 3D grounding as a multi-turn GUI interaction: the current state is rendered into a graphical observation, a vision-language model (VLM) predicts one executable edit, and the edited state is re-rendered for the next turn. The rendering pairs the scene point cloud with the interface elements that make an edit legible, namely the 3D model overlay, its 3D box, and a gizmo exposing the object's local frame, in which every edit is expressed.

\begin{figure}[t]
    \centering
    \includegraphics[width=\linewidth]{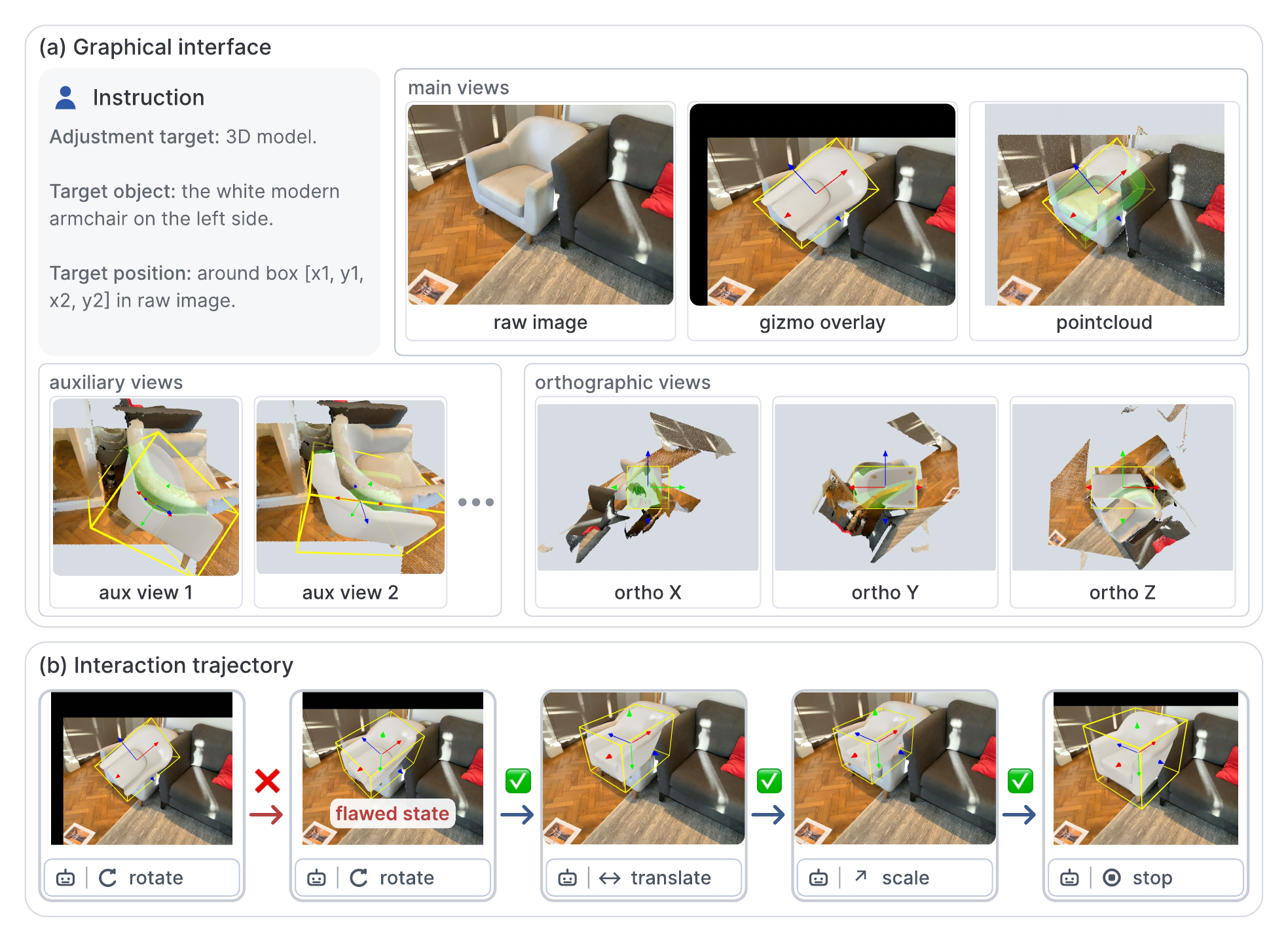}
    \caption{\textbf{GizmoAct graphical observation and interaction trajectory.} (a) The observation at one turn. The instruction block specifies the adjustment target and the referring cue; the main views pair the raw input image with renderings that overlay the asset, its yellow 3D box, and the gizmo showing the object's local frame (red/green/blue axes), in which all edits are expressed. Semi-transparent green marks the model surface occluded by the scene point cloud. Auxiliary views add multi-view evidence, and orthographic views along the local axes support fine refinement. (b) A rollout in which the policy issues a flawed edit (red cross) and recovers from the resulting state (green checks). During SFT data construction, we simulate such cases through error injection and supervise only the recovery actions.}
    \label{fig:gizmoact-observation}
\end{figure}

We instantiate this formulation for the 9-DoF object pose estimation required by composable scene modeling. The manipulated element for object $o$ is the 3D asset $A_o$ generated in Sec.~\ref{sec:amodal-generation}, and its state
\begin{equation}
    x_t = (p_t, R_t, s_t), \qquad
    p_t \in \mathbb{R}^3,\quad R_t \in SO(3),\quad s_t \in \mathbb{R}_{+}^{3},
    \label{eq:gizmoact-state}
\end{equation}
collects the object center $p_t$, the object-to-world rotation $R_t$, and the anisotropic object scale $s_t$. Everything else comes from the evidence bundle $\mathcal{E}_o$ of Sec.~\ref{sec:evidence}: the target cue $c_o$ (referring text and position) identifies the object, the reference views and the point cloud $\mathcal{P}_o$ are rendered into the observation of \Cref{fig:gizmoact-observation}(a), and an initial object pose provides $x_0$. A rollout can start from an arbitrary $x_0$; in \ourwork we take it from the coarse layout with arbitrary orientation. One turn then consists of a rendering, an action, and a state update:
\begin{equation}
    o_t = \mathcal{R}\!\left(A_o, x_t, \mathcal{E}_o, v_t\right), \quad
    a_t \sim \pi_\theta(\cdot \mid o_{\leq t}, a_{<t}), \quad
    (x_{t+1}, v_{t+1}) = F(x_t, a_t),
    \label{eq:gizmoact-loop}
\end{equation}
where $v_t$ selects the rendering mode of the turn, which an action can change without touching the object state. The policy $\pi_\theta$ is a VLM: it reads the renderings as images and emits $a_t$ as text, so a rollout is a single multi-turn conversation. As \Cref{fig:gizmoact-observation}(b) shows, it sees the rendered consequence of its own edits, which might be flawed, and keeps updating until it considers the proxy aligned and predicts a stop action.

\subsubsection{Graphical Observation Interface}

\paragraph{Graphical elements.}
Beyond the image and point-cloud renderings, an observation contains three elements. The rendered 3D model indicates the current object pose. The 3D box provides clearer object boundary information and helps the VLM judge the object size. Most importantly, the 3D gizmo defines the local coordinate frame of the object, and all pose updates are predicted relative to this frame.

\paragraph{Viewpoints.}
We provide three complementary viewpoints. Beyond the main views, the auxiliary views supply additional multi-view evidence, and the orthographic views are rendered along the local axes under the current object pose, which helps fine pose refinement. The orthographic views also act as pseudo multi-view observations for a single-view input, resolving the scale-depth ambiguity that leaves translation and scale under-determined from a single viewpoint.

\paragraph{Occlusion cue.}
A single point-cloud rendering often leaves the depth relation between the 3D model and the point cloud ambiguous. We alleviate this by overlaying an additional semi-transparent green layer on the occluded regions of the model, namely the model surface that falls behind the point cloud from the current viewpoint.

\subsubsection{GizmoAct Action Space}

The core action space of GizmoAct contains only two actions: \texttt{update\_pose}, which edits the object pose, and \texttt{stop}, which ends the episode. Auxiliary actions that further facilitate pose updates can also be added.

\paragraph{Continuous pose update.}
\texttt{update\_pose} rests on two design choices. First, the gizmo defines the local frame of the object, and every update is an incremental edit expressed in that frame, so the policy never predicts an absolute pose, which is more ambiguous to predict from images. Second, translation and scale are predicted relative to the current object size rather than as metric quantities, which removes the dependence on metric scale estimation. Concretely, given a ZXY Euler-angle delta $\delta_r=(\delta_z,\delta_x,\delta_y)\in\mathbb{R}^3$ in degrees, a translation delta $\delta_p\in\mathbb{R}^3$, and a relative scale delta $\delta_s\in\mathbb{R}^3$,
\begin{equation}
    R_{t+1} = R_t R_{\mathrm{ZXY}}(\delta_r), \qquad
    s_{t+1} = s_t \odot (\mathbf{1} + \delta_s), \qquad
    p_{t+1} = p_t + R_{t+1}(\delta_p \odot s_t).
    \label{eq:gizmoact-pose-update}
\end{equation}
Rotation is applied first, so $\delta_p$ acts in the frame that the next gizmo will display, and both $\delta_p$ and $\delta_s$ are measured in units of the current size $s_t$. The VLM predicts \texttt{stop} when it considers the current object pose accurate enough.

\begin{figure}[t]
    \centering
    \includegraphics[width=0.88\linewidth]{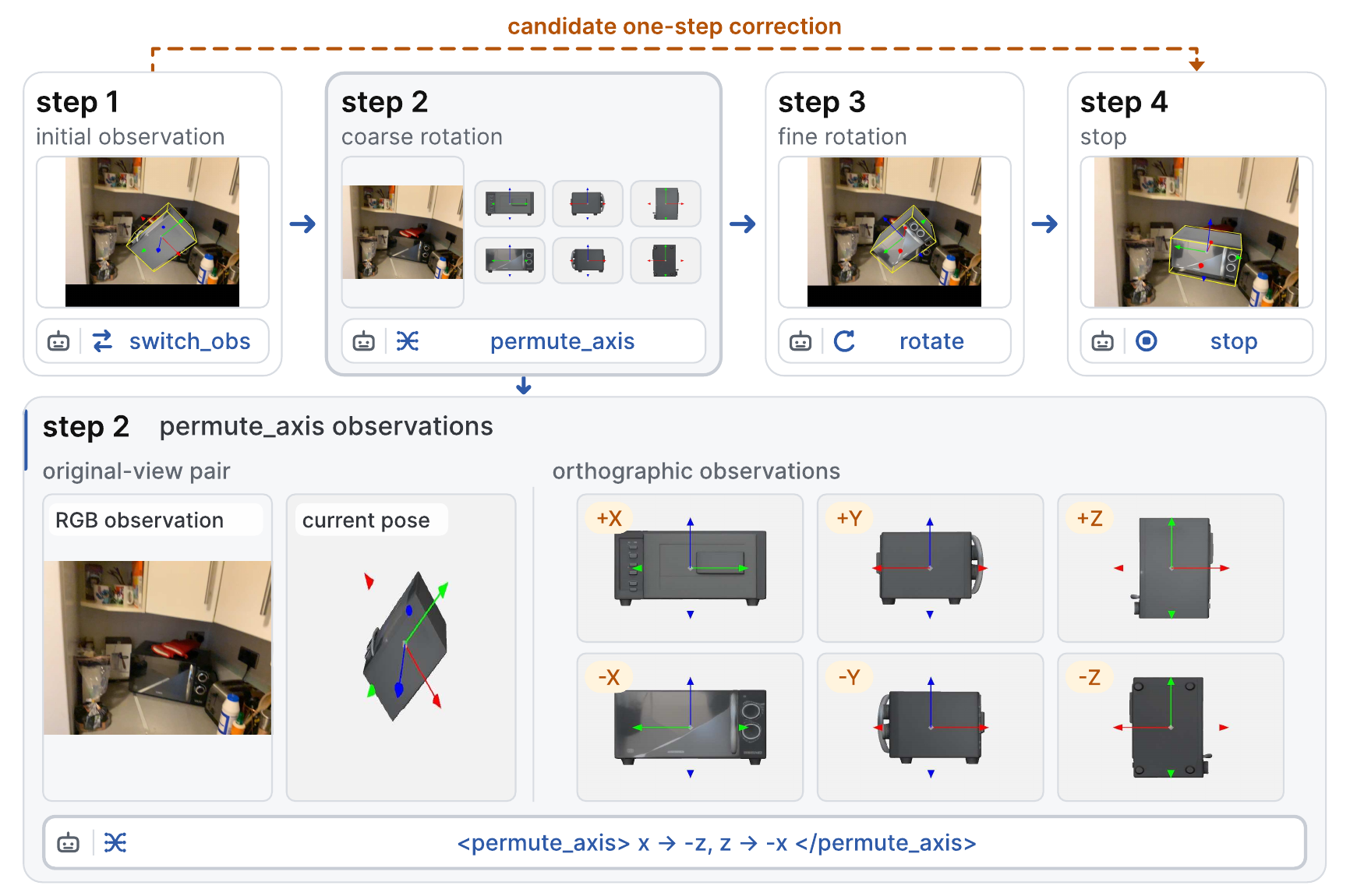}
    \caption{\textbf{Extended actions for coarse rotation.} When the initial rotation error is large (step 1), the policy first issues \texttt{switch\_obs}, which changes only the rendering mode: the next observation (step 2, enlarged in the bottom panel) shows the original-view pair together with six orthographic renderings of the asset along its signed local axes, exposing the sides the main views hide. From these views the policy predicts a single \texttt{permute\_axis} action, which selects one of the 24 axis-aligned reorientations and removes the dominant rotation residual in one step. The remaining error is repaired by \texttt{update\_pose} (step 3) before the policy predicts \texttt{stop} (step 4). The dashed arrow marks the alternative of predicting the whole correction with a single regular \texttt{update\_pose}, skipping the permutation, which is less accurate.}
    \label{fig:gizmoact-action-trajectory}
\end{figure}

\paragraph{Extended actions.}
A large initial rotation error is difficult to resolve with \texttt{update\_pose} alone. The main views expose only the object surface facing the input camera, which rarely reveals how a generated asset is oriented, and correcting a large rotation residual with Euler deltas risks gimbal lock. GizmoAct therefore adds two actions for coarse rotation, illustrated in \Cref{fig:gizmoact-action-trajectory}. \texttt{switch\_obs} leaves $x_t$ untouched and re-renders the asset along six signed axes, exposing the sides the main views hide. From these views, the policy emits an axis permutation with \texttt{permute\_axis}, which names the target $x$ and $z$ axes with signed source axes $e_x, e_z \in \{\pm\hat{x},\pm\hat{y},\pm\hat{z}\}$ satisfying $e_x^\top e_z = 0$ and derives $y$ by the right-hand rule:
\begin{equation}
    M(e_x, e_z) = [\,e_x,\; e_z \times e_x,\; e_z\,], \qquad
    R_{t+1} = R_t M(e_x, e_z).
    \label{eq:gizmoact-axis-remap}
\end{equation}
The $6\times 4$ valid $(e_x, e_z)$ pairs cover all 24 axis-aligned reorientations, so a coarse rotation costs a single action and \texttt{update\_pose} is left with the residual.

\paragraph{Executable syntax.}
Every action is written as one XML-style tag whose body is a JSON payload, and the VLM emits exactly one tag per turn:
\begin{actionbox}
\begin{verbatim}
t=1   <switch_obs>permute_axis</switch_obs>
t=2   <permute_axis>{"x":"-y","z":"x"}</permute_axis>
t=3   <update_pose>{"rotate":{"z":8.25,"x":-2.50,"y":3.10,"order":"ZXY"},
                    "translate":{"x":0.10,"y":-0.04,"z":0.02},
                    "scale":{"x":0.05,"y":-0.03,"z":0.02}}</update_pose>
      ...
t=T   <stop>{}</stop>
\end{verbatim}
\end{actionbox}
An \texttt{update\_pose} payload may carry any non-empty subset of the rotation, translation, and scale fields. Malformed payloads, empty bodies, and unknown action names are rejected by the environment.

\subsubsection{Supervised Finetuning on Synthetic Trajectories}

We finetune a pretrained VLM into the GizmoAct policy on synthetic expert trajectories. Each training example pairs the evidence bundle $\mathcal{E}_o$ with a generated or provided object asset, an initial pose $x_0$ drawn from a perturbed pose distribution, and a hidden target pose $x^\star$ available only to the trajectory generator. Rolling the example out in the same executable environment used at inference time yields the rendered observations and XML actions, which we pack as a multi-turn conversation.

\paragraph{Data mixture.}

We combine three complementary data sources. \emph{Populated 3D-FRONT} places MesaTask~\citep{hao2025mesatask} tabletop layouts onto support surfaces in 3D-FRONT~\citep{fu20213dfront} scenes, giving clean indoor geometry and dense tabletop arrangements. \emph{FoundationPose}~\citep{wen2023foundationpose} scatters retextured assets on textured planes under randomized lighting, stressing arbitrary orientation, heavy occlusion, and appearance mismatch between the asset and the image. \emph{CA-1M Objects} samples long-tail categories from the train split of CA-1M~\citep{lazarow2024cubify}, following its original train/val partition, generates assets with SAM~3D~\citep{sam3dteam2025sam3d}, and manually aligns them to CA-1M LiDAR, contributing noisy real point clouds and generated assets whose geometry ranges from accurate to visibly mismatched with the observation.

\paragraph{Privileged expert policy.}
Trajectories follow a preferred repair order: rotation first, then translation and scale by residual magnitude, with one action allowed to cover several dimensions. The generator has oracle access to $x^\star$ and tracks rotation error, translation error normalized by the target diagonal, and log-scale error. A dimension stays active while its residual exceeds the two-decimal precision of the interface, and the expert emits \texttt{stop} once none is active. Before any \texttt{update\_pose} turn, it checks for a coarse rotation fix: when a signed-axis permutation removes a large and unambiguous part of the rotation residual, it requests the \texttt{permute\_axis} view with \texttt{switch\_obs} and applies the best of the 24 permutations. The rest is repaired by \texttt{update\_pose} actions that each fully cancel the error on a sampled subset of the active dimensions, and sampling the subset and the turn count diversifies trajectory length.

\paragraph{Error injection and recovery.}
Expert trajectories alone never show the policy how to recover from its own errors. Following the noise-injected demonstrations of DART~\citep{laskey2017dart}, we inject errors into the expert rollout, illustrated in \Cref{fig:gizmoact-observation}(b): corrupted \texttt{update\_pose} commands (under-correction, overshoot, wrong axis, etc.) and wrong or skipped axis permutations. A candidate injected error $\tilde{a}_t$ is kept only if a clean expert, simulated from the state it produces, still reaches the target within the remaining turn budget. The kept action is executed and stays in the context as history, and the expert then demonstrates recovery from the resulting state. A trajectory carries a few such injections, spread across turns and error dimensions. Formally, for target token $y_{t, j}^{\star}$ under history $h_t=(o_{\leq t}, a_{<t})$, the SFT objective is
\begin{equation}
    \mathcal{L}_{\mathrm{SFT}}(\theta)
    =
    -
    \mathbb{E}_{\xi \sim \mathcal{D}_{\mathrm{SFT}}}
    \sum_{t}\sum_{j}
    w_t m_{t, j}
    \log \pi_\theta
    \left(y_{t, j}^{\star} \mid h_t, y_{t, <j}^{\star}\right),
    \label{eq:gizmoact-sft}
\end{equation}
where $m_{t, j}$ selects the action tokens emitted at turn $t$ and $w_t \in \{0, 1\}$ masks out the injected errors.

\subsubsection{RL Training}
The finetuned VLM policy is limited in three ways. Its rollout success rate is still low on difficult instances; it recovers poorly from the states its own errors produce, because injected errors are scripted and cover only a part of the on-policy contexts the policy actually reaches; and it falls short of adapting the correction schedule to what it can perceive. We therefore continue training with GRPO \citep{shao2024deepseekmath} on online rollouts, in the same executable environment used at inference. We elaborate on the reward design and training scheme below.

\paragraph{Reward design.}
A rollout is scored only at its final state, and that scalar is broadcast to every action token of the trajectory. We score two axes: the generalized 3D IoU $g$ between the predicted and the target oriented 3D box, which serves as a proxy for translation and scale error, and the geodesic rotation error $e_R$. We do not score ADD-S \citep{xiang2017posecnn}, which assumes symmetry annotations that cover only part of our data; ADD-SB drops the reliance but is blind to texture, hence to orientations that only appearance distinguishes. Instead of mapping the two axes to a continuous score, we quantize each into four levels, bad, coarse, success, and excellent, with boundaries at $g=0.75, 0.85, 0.925$ and $e_R=30^\circ, 10^\circ, 5^\circ$. Writing $\ell_g, \ell_R \in \{0,1,2,3\}$ for the two levels and $S=\mathbf{1}[\ell_g \geq 2] \cdot \mathbf{1}[\ell_R \geq 2]$ for joint success, the reward of a trajectory $\xi$ with final action $a_T$ is
\begin{equation}
    r(\xi)
    =
    \frac{S}{8}
    \left(
        4
        + 2 \cdot \mathbf{1}[\ell_g=3]
        + 2 \cdot \mathbf{1}[\ell_R=3]
        + \mathbf{1}[a_T=\text{\texttt{stop}}]
    \right).
    \label{eq:gizmoact-reward}
\end{equation}
A trajectory earns nothing until both axes reach the success band, gains extra reward for each axis that reaches excellent, and receives a final bonus for stopping once aligned. Rollouts that emit an invalid action are scored zero. Since $e_R$ penalizes symmetry-equivalent poses, we exclude symmetric objects from RL training; scoring them properly would require symmetry annotations in the reward. Quantization keeps the group comparison informative and avoids noisy advantages introduced by continuous scores.

\paragraph{Dynamic sampling.}
We follow DAPO \citep{yu2025dapo} and use dynamic sampling to keep each batch informative. For each prompt we draw $K=8$ rollouts at temperature $0.7$ and reject the group in two cases: when any rollout fails to complete, and when the $K$ terminal rewards all fall on the same level. The second case covers uniformly failed, uniformly solved, and tied groups, none of which carries a ranking signal. Rejected groups are discarded and sampling continues until the batch is filled.

\paragraph{Policy optimization.}
Apart from the reward and the group filter above, we follow an established GRPO training setup. We adopt the token-level clipped objective of DAPO \citep{yu2025dapo} with asymmetric clipping and dual clip. We remove the group standard-deviation normalization of the advantage, as in Dr.GRPO \citep{liu2025understanding}. We include truncated importance sampling (TIS) \citep{yao2025offpolicy} to compensate for the rollout--training mismatch. We disable KL penalty in optimization. Each iteration collects at least $40$ prompts and $320$ trajectories, slightly more when dynamic sampling refills the batch, and every collected batch is optimized for two epochs.

\subsection{Scene Composition and Postprocessing}
\label{sec:postprocess}

After individual objects are generated and grounded, \ourwork composes them into a full indoor scene. The base output is an object-centric scene graph with object assets, poses, scales, categories, and spatial relations. Optional postprocessing can refine support relations, collision, contact consistency, and placement plausibility. These operations are intentionally separated from GizmoAct: the agent focuses on grounding individual editable components, while postprocessing handles scene-level consistency constraints that are easier to express as rules, physics checks, or global verification.

\section{Experiments}

\begin{figure}[t]
    \centering
    \includegraphics[width=0.82\linewidth]{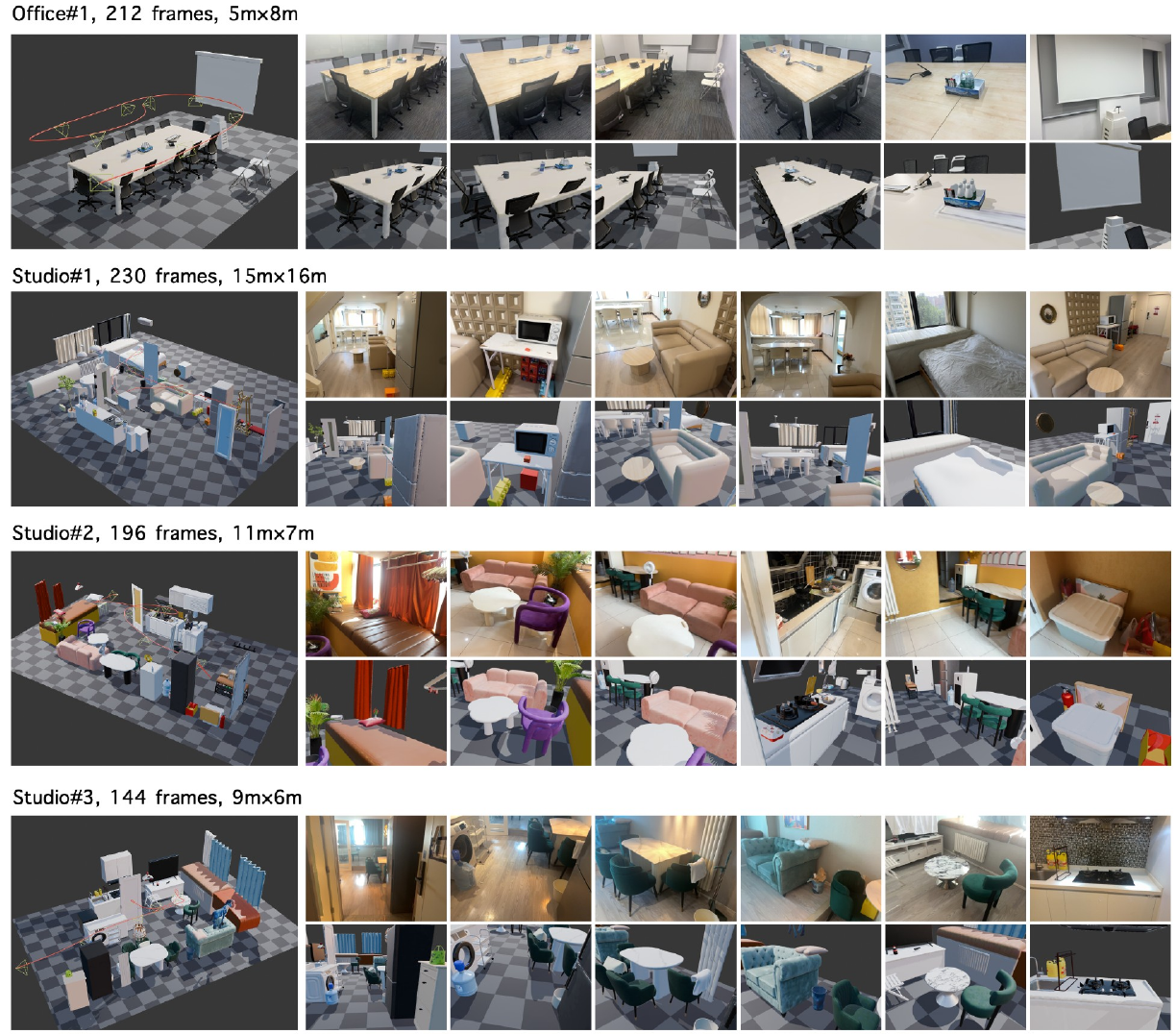}
    \caption{\textbf{Gallery of composed scenes.} Four real captures reconstructed by \ourwork, annotated with the number of captured frames and the room extent. For each scene, the left panel shows the composed object-centric scene with the capture trajectory and camera frusta overlaid; the right panels pair captured frames (top row) with renderings of the reconstructed scene from the same camera poses (bottom row). Every object is an individual complete asset, so the composed scenes are directly editable and simulation-ready.}
    \label{fig:gallery}
\end{figure}

\noindent We evaluate \ourwork at three levels.
Section~\ref{sec:exp-video-layout} evaluates scene-level 3D object detection, measuring object recovery and localization by the scene-parsing module.
Section~\ref{sec:exp-object-pose} evaluates object pose estimation, measuring the alignment accuracy of GizmoAct independently of upstream detection.
Section~\ref{sec:exp-scene-reconstruction} evaluates the complete system through scene reconstruction, jointly measuring reconstructed object geometry and final scene layout.
Section~\ref{sec:exp-ablation} ablates the scene-parsing components and GizmoAct training strategies.

\subsection{Scene-Level 3D Object Detection}
\label{sec:exp-video-layout}

This experiment evaluates the scene-level 3D object boxes recovered by the
multi-view object-centric scene parsing module in Sec.~\ref{sec:evidence}. It
measures whether the parser recovers the target object instances and localizes
them with 3D boxes before downstream asset generation and GizmoAct pose
refinement. We compare its scene-level output with the per-scene predictions
produced by the baseline pipelines.

\paragraph{Baselines.}
We compare against Boxer~\citep{boxer2026} and WildDet3D~\citep{wilddet3d2026},
which recover 3D object boxes from image observations and external object
prompts. To control which objects are queried, all methods receive prompts for
the same target instances. We evaluate two prompt-frame settings. The
\emph{all} setting provides prompts on every annotated frame, whereas the
\emph{key} setting provides them only on the geometry-aware keyframes selected
by our method. We evaluate Boxer in both settings and WildDet3D in the key
setting. Both baselines use the offline fusion implementation in
Boxer. Our method starts from the keyframe prompts and then runs the complete
scene-parsing pipeline of Sec.~\ref{sec:evidence}, including object-centric
full-sequence evidence consolidation and relation-aware scene refinement.

\paragraph{Datasets and metrics.}
We evaluate on CA-1M~\citep{lazarow2024cubify} and R2S-Scene. CA-1M provides
independently laser-scanned geometry and near-exhaustive object annotations that
include long-tail categories and small and thin instances across captured
sequences. R2S-Scene, the scene-level split of our real-world real-to-sim
benchmark, contains verified 3D boxes and poses for the objects included in
scene reconstruction. The $\_\mathrm{all}$ protocol evaluates all annotated
ground-truth boxes. The $\_\mathrm{filter}$ protocol evaluates ground-truth
objects recovered by at least one compared method, excluding objects missed by
every method. For each scene, we compute class-agnostic AP at 3D IoU thresholds
from 0.05 to 0.50 in increments of 0.05, average over thresholds, and report the
mean across scenes.

\paragraph{Results.}
\Cref{tab:object-grounding} shows that \ourwork achieves the highest mAP in all
four protocols. On CA-1M, it improves the strongest baseline from 0.171 to
0.180 under the $\_\mathrm{all}$ protocol and from 0.373 to 0.390 under the
$\_\mathrm{filter}$ protocol. On R2S-Scene, mAP increases from 0.351 to 0.592
under the $\_\mathrm{all}$ protocol and from 0.355 to 0.597 under the
$\_\mathrm{filter}$ protocol. Although Boxer improves when given prompts on all
frames, our method, initialized only from keyframe prompts, still performs
better while using the full sequence for instance propagation, multi-view
validation, and 3D box refinement. This comparison highlights the benefit of
organizing full-sequence observations around object hypotheses rather than
relying only on direct fusion of independently predicted per-frame boxes.
Sec.~\ref{sec:exp-ablation} further evaluates object-centric full-sequence evidence consolidation
and relation-aware scene refinement.

\begin{table}[t]
    \caption{\textbf{Scene-level 3D object detection on CA-1M and R2S-Scene.} Per-scene entries report class-agnostic mAP, and Mean averages across scenes. Prompt frames \emph{all}/\emph{key} denotes prompts on all annotated frames or selected keyframes; \_all/\_filter denotes evaluation on all annotations or only objects recovered by at least one method. A dash denotes no valid prediction. Best mean in bold.}
    \label{tab:object-grounding}
    \centering
    \scriptsize
    \setlength{\tabcolsep}{3pt}
    \renewcommand{\arraystretch}{1.08}

    \textbf{CA-1M\_all}
    \resizebox{\textwidth}{!}{%
    \begin{tabular}{@{}llccccccccccc@{}}
        \toprule
        Method & Prompt frames & 01 & 02 & 03 & 04 & 05 & 06 & 07 & 08 & 09 & 10 & Mean \\
        \midrule
        Boxer & all & 0.151 & 0.142 & 0.224 & 0.134 & 0.241 & 0.106 & 0.170 & 0.302 & 0.113 & 0.122 & 0.171 \\
        Boxer & key & 0.037 & 0.027 & 0.138 & 0.057 & 0.059 & 0.019 & 0.095 & 0.122 & -- & 0.053 & 0.067 \\
        WildDet3D & key & 0.051 & 0.014 & 0.101 & 0.043 & 0.113 & 0.035 & 0.025 & 0.088 & 0.036 & 0.065 & 0.057 \\
        Ours & key & 0.168 & 0.148 & 0.268 & 0.138 & 0.266 & 0.109 & 0.170 & 0.319 & 0.105 & 0.110 & \textbf{0.180} \\
        \bottomrule
    \end{tabular}}

    \vspace{0.6em}
    \textbf{CA-1M\_filter}
    \resizebox{\textwidth}{!}{%
    \begin{tabular}{@{}llccccccccccc@{}}
        \toprule
        Method & Prompt frames & 01 & 02 & 03 & 04 & 05 & 06 & 07 & 08 & 09 & 10 & Mean \\
        \midrule
        Boxer & all & 0.312 & 0.344 & 0.383 & 0.339 & 0.544 & 0.264 & 0.423 & 0.404 & 0.404 & 0.314 & 0.373 \\
        Boxer & key & 0.064 & 0.051 & 0.242 & 0.134 & 0.119 & 0.038 & 0.237 & 0.156 & 0.119 & 0.131 & 0.129 \\
        WildDet3D & key & 0.092 & 0.024 & 0.170 & 0.103 & 0.247 & 0.084 & 0.063 & 0.114 & 0.119 & 0.164 & 0.118 \\
        Ours & key & 0.349 & 0.358 & 0.461 & 0.348 & 0.603 & 0.273 & 0.428 & 0.426 & 0.378 & 0.279 & \textbf{0.390} \\
        \bottomrule
    \end{tabular}}

    \vspace{0.6em}
    \textbf{R2S-Scene\_all}
    \resizebox{\textwidth}{!}{%
    \begin{tabular}{@{}llcccccccccc@{}}
        \toprule
        Method & Prompt frames & 01 & 02 & 03 & 04 & 05 & 06 & 07 & 08 & 09 & Mean \\
        \midrule
        Boxer & all & 0.458 & 0.728 & 0.240 & 0.308 & 0.333 & 0.090 & 0.309 & 0.386 & 0.306 & 0.351 \\
        Boxer & key & 0.172 & 0.524 & 0.048 & 0.078 & 0.089 & 0.084 & 0.059 & 0.122 & 0.132 & 0.145 \\
        WildDet3D & key & 0.148 & 0.309 & 0.080 & -- & 0.080 & 0.024 & 0.123 & 0.193 & 0.043 & 0.125 \\
        Ours & key & 0.613 & 0.959 & 0.507 & 0.379 & 0.523 & 0.370 & 0.596 & 0.781 & 0.603 & \textbf{0.592} \\
        \bottomrule
    \end{tabular}}

    \vspace{0.6em}
    \textbf{R2S-Scene\_filter}
    \resizebox{\textwidth}{!}{%
    \begin{tabular}{@{}llcccccccccc@{}}
        \toprule
        Method & Prompt frames & 01 & 02 & 03 & 04 & 05 & 06 & 07 & 08 & 09 & Mean \\
        \midrule
        Boxer & all & 0.471 & 0.728 & 0.241 & 0.311 & 0.341 & 0.090 & 0.309 & 0.401 & 0.306 & 0.355 \\
        Boxer & key & 0.181 & 0.524 & 0.048 & 0.083 & 0.091 & 0.084 & 0.059 & 0.124 & 0.132 & 0.147 \\
        WildDet3D & key & 0.153 & 0.309 & 0.082 & -- & 0.082 & 0.024 & 0.123 & 0.202 & 0.043 & 0.127 \\
        Ours & key & 0.632 & 0.959 & 0.512 & 0.383 & 0.537 & 0.370 & 0.593 & 0.788 & 0.603 & \textbf{0.597} \\
        \bottomrule
    \end{tabular}}
\end{table}

\subsection{Layout Refinement and Object Pose Estimation}
\label{sec:exp-object-pose}

\begin{table}[t!]
    \caption{\textbf{Object pose estimation on R2S-Object, CA-1M, and ADT.} ADD-SB and its thresholded success rates measure surface alignment, and 3D IoU measures oriented-box overlap. Any6D and GizmoAct receive the same generated asset for each sample, whereas SAM~3D and RecGen jointly predict shape and pose; GizmoAct refines from Boxer initialization, and the max-4-view variant uses up to four valid views. GizmoAct improves strict alignment (ADD-SB@0.05) and 3D IoU over all baselines on the three datasets. Best per dataset in bold.}
    \label{tab:object-pose-estimation}
    \centering
    \resizebox{0.92\linewidth}{!}{%
    \begin{tabular}{llcccc}
        \toprule
        Dataset & Method & ADD-SB $\downarrow$ & ADD-SB@0.1 $\uparrow$ & ADD-SB@0.05 $\uparrow$ & 3D IoU $\uparrow$ \\
        \midrule
        \multirow{6}{*}{R2S-Object}
        & Any6D & 0.086 & 76.6\% & 46.8\% & 0.388 \\
        & SAM 3D & 0.050 & 92.8\% & 61.6\% & 0.482 \\
        & RecGen (1 view) & 0.040 & 98.4\% & 79.2\% & 0.499 \\
        & RecGen (2 views) & 0.038 & 96.8\% & 77.6\% & 0.500 \\
        \cmidrule(lr){2-6}
        & GizmoAct (1 view) & 0.019 & 98.4\% & 88.0\% & 0.685 \\
        & GizmoAct (max 4 views) & \textbf{0.017} & \textbf{99.2\%} & \textbf{92.0\%} & \textbf{0.719} \\
        \midrule
        \multirow{6}{*}{CA-1M}
        & Any6D & 0.116 & 67.3\% & 42.2\% & 0.355 \\
        & SAM 3D & 0.062 & 86.4\% & 56.3\% & 0.434 \\
        & RecGen (1 view) & 0.061 & 87.9\% & 57.8\% & 0.391 \\
        & RecGen (2 views) & 0.046 & 88.4\% & 57.3\% & 0.400 \\
        \cmidrule(lr){2-6}
        & GizmoAct (1 view) & 0.023 & 94.0\% & 81.9\% & 0.600 \\
        & GizmoAct (max 4 views) & \textbf{0.021} & \textbf{96.0\%} & \textbf{83.4\%} & \textbf{0.607} \\
        \midrule
        \multirow{6}{*}{ADT}
        & Any6D & 0.047 & 85.0\% & 66.7\% & 0.520 \\
        & SAM 3D & 0.063 & 90.0\% & 41.7\% & 0.404 \\
        & RecGen (1 view) & 0.044 & 96.7\% & 71.7\% & 0.462 \\
        & RecGen (2 views) & 0.039 & 95.0\% & 70.0\% & 0.444 \\
        \cmidrule(lr){2-6}
        & GizmoAct (1 view) & \textbf{0.020} & \textbf{98.3\%} & 86.7\% & 0.664 \\
        & GizmoAct (max 4 views) & 0.021 & 91.7\% & \textbf{90.0\%} & \textbf{0.670} \\
        \bottomrule
    \end{tabular}}
\end{table}

\paragraph{Setup.}
We evaluate GizmoAct independently of the upstream scene-level detection module so that global layout errors do not confound pose refinement. Each task provides target cues (2D bounding box and object category), posed RGB-D inputs, 3D object model, and a coarse object pose. GizmoAct supports Boxer~\citep{boxer2026} initialization, a depth-and-mask initialization following Any6D~\citep{lee2025any6d}, and initialization from SAM~3D~\citep{sam3dteam2025sam3d}. The main comparison uses Boxer initialization with a policy whose RL stage is trained on Boxer-initialized poses (this specialization is ablated in Sec.~\ref{sec:exp-ablation}), and reports a single-view variant and a multi-view variant that receives at most four valid views. Both variants refine the 9-DoF object pose for at most 12 steps.

\paragraph{Datasets.}
We evaluate on CA-1M~\citep{lazarow2024cubify}, Aria Digital Twin (ADT)~\cite{pan2023aria}, and the object split of our R2S benchmark, denoted R2S-Object. The CA-1M subset contains 199 objects from 125 categories in the CA-1M validation set. We associate each object across views, estimate and manually filter its masks with SAM~3~\citep{carion2026sam3segmentconcepts}, generate its 3D model with SAM~3D~\citep{sam3dteam2025sam3d}, and manually annotate a 9-DoF pose that aligns the model with the LiDAR point cloud. We form the evaluation depth maps by completing the projected FARO LiDAR points with LingBot-Depth~\citep{tan2026maskeddepthmodelingspatial}. R2S-Object contains 125 objects from 64 categories captured in real indoor scenes. We initialize their layouts with the scene-level detection pipeline from Sec.~\ref{sec:exp-video-layout}, generate object models with Seed3D~2.0~\citep{gu2026seed3d20advancinghighfidelity}, and apply the same mask filtering and manual pose annotation protocol as CA-1M. Pi3X~\citep{wang2026pi3permutationequivariantvisualgeometry} provides the R2S-Object depth estimates. From ADT~\citep{pan2023aria}, we select 60 objects with designer-authored meshes and accurate object poses; we render their depth maps from the ground-truth scene geometry.

\paragraph{Metrics.}
Following SAM~3D~\cite{sam3dteam2025sam3d} and RecGen~\citep{zadaianchuk2026recgen}, we measure the bidirectional average nearest-neighbor distance between the predicted and ground-truth posed surfaces. We report this distance as ADD-SB in meters. ADD-SB@0.1 and ADD-SB@0.05 are the percentages of predictions whose ADD-SB falls below 10\% and 5\% of the ground-truth object diameter. We also report 3D IoU between the oriented bounding boxes of the predicted and ground-truth posed objects.

\paragraph{Baselines.}
The comparison covers two method families. Any6D~\citep{lee2025any6d} estimates the scale and 6-DoF pose of an independently generated object model from a single RGB-D observation. Within each split, Any6D and GizmoAct receive the same generated asset. SAM~3D and RecGen jointly estimate object shape and 9-DoF pose from RGB or RGB-D observations and a target mask. We evaluate the single-view and two-view variants of RecGen, together with the single-view and max-4-view variants of GizmoAct.

\paragraph{Results.}
\Cref{tab:object-pose-estimation} shows that GizmoAct improves strict alignment and oriented-box overlap on all three datasets. On R2S-Object, the max-4-view variant reduces ADD-SB from 0.038 for the strongest baseline to 0.017, raises ADD-SB@0.05 from 79.2\% to 92.0\%, and improves 3D IoU from 0.500 to 0.719. On CA-1M, it reduces ADD-SB from 0.046 to 0.021 while increasing ADD-SB@0.05 from 57.8\% to 83.4\% and 3D IoU from 0.434 to 0.607. The ADT results separate the benefits of the two input settings: the single-view variant obtains the lowest ADD-SB (0.020) and the highest ADD-SB@0.1 (98.3\%), while the max-4-view variant gives the best ADD-SB@0.05 (90.0\%) and 3D IoU (0.670).

\begin{table}[t!]
    \caption{\textbf{Robustness to pose initialization.} All rows evaluate the same GizmoAct policy, trained with randomly perturbed initial poses, using at most four views and 12 refinement steps; only the initial pose differs. Any6D* denotes the depth-and-mask initialization from Any6D without FoundationPose prediction or refinement. Boxer initialization works best with the estimated depth of R2S-Object and CA-1M, whereas the exact geometry of ADT favors the Any6D* and SAM~3D initializers.}
    \label{tab:object-pose-initialization}
    \centering
    \resizebox{\linewidth}{!}{%
    \begin{tabular}{lccccccccc}
        \toprule
        & \multicolumn{3}{c}{R2S-Object}
        & \multicolumn{3}{c}{CA-1M}
        & \multicolumn{3}{c}{ADT} \\
        \cmidrule(lr){2-4}\cmidrule(lr){5-7}\cmidrule(lr){8-10}
        Initialization & ADD-SB $\downarrow$ & ADD-SB@0.05 $\uparrow$ & 3D IoU $\uparrow$
        & ADD-SB $\downarrow$ & ADD-SB@0.05 $\uparrow$ & 3D IoU $\uparrow$
        & ADD-SB $\downarrow$ & ADD-SB@0.05 $\uparrow$ & 3D IoU $\uparrow$ \\
        \midrule
        Boxer & \textbf{0.022} & 90.4\% & \textbf{0.678} & \textbf{0.026} & \textbf{77.9\%} & \textbf{0.568} & 0.025 & 81.7\% & 0.626 \\
        Any6D* & 0.030 & 84.8\% & 0.612 & 0.028 & 73.9\% & 0.518 & \textbf{0.016} & 93.3\% & 0.641 \\
        SAM 3D & 0.025 & \textbf{91.2\%} & 0.633 & 0.028 & 77.4\% & 0.542 & 0.022 & \textbf{95.0\%} & \textbf{0.678} \\
        \bottomrule
    \end{tabular}}
\end{table}

\paragraph{Robustness to pose initialization.}
We test whether one GizmoAct policy adapts to different pose initializations without retraining. This experiment uses the general policy trained with randomly perturbed initial poses, rather than the Boxer-specialized policy of the main comparison; it receives at most four views for up to 12 refinement steps. Boxer combines a randomly sampled local rotation with its predicted 3D box, which typically leaves a large rotation error but smaller translation and scale errors. The Any6D-style initializer estimates the initial pose from the target depth and instance mask, so its errors depend on object geometry and viewpoint and can affect rotation, translation, and scale jointly. SAM~3D usually leaves only local residual errors. \Cref{tab:object-pose-initialization} shows that Boxer performs best overall with noisy estimated depth on R2S-Object and CA-1M, whereas the exact depth and camera poses of ADT favor the Any6D-style and SAM~3D initializers. The policy therefore adapts to heterogeneous initialization errors, although the best initializer still depends on the accuracy of the input geometry.

\begin{figure}[p]
    \centering
    \includegraphics[width=\linewidth,height=\dimexpr\textheight-7\baselineskip\relax,keepaspectratio]{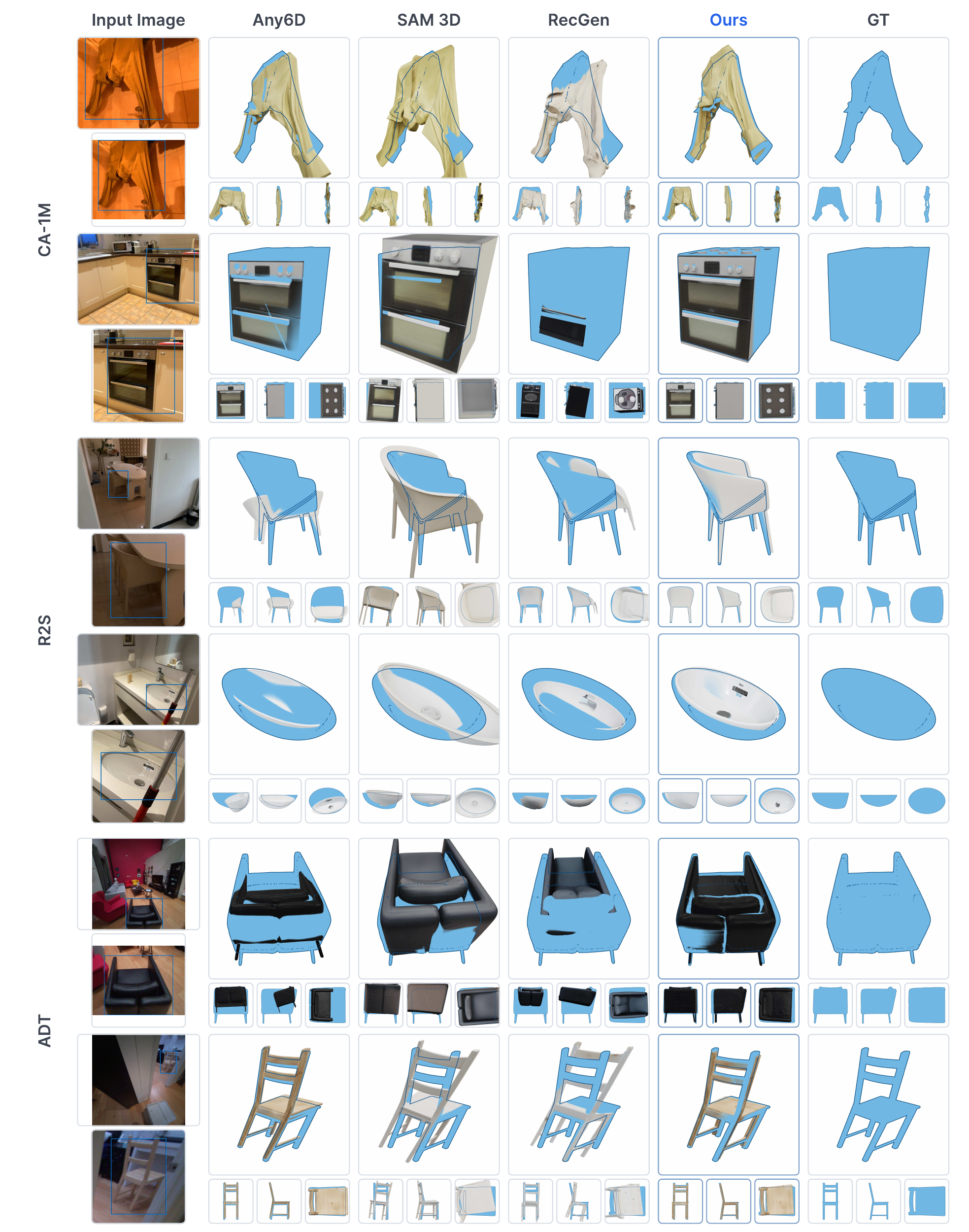}
    \caption{\textbf{Qualitative comparison of object pose estimation on CA-1M, R2S-Object, and ADT.} The leftmost column shows the scene image and a close-up with the target instance boxed. Blue renders the ground-truth model at the ground-truth pose; each method's model, posed by its prediction, is rendered jointly with this blue reference, so their mutual occlusion---together with the three axis-aligned close-ups---reveals the pose accuracy, and the GT column shows the reference alone. \ourwork aligns the generated assets more closely with the observed instances across categories, shape complexity, and viewpoints.}
    \label{fig:qualitative-compare-rank-1-2}
\end{figure}

\begin{figure}[p]
    \centering
    \includegraphics[width=\linewidth,height=\dimexpr\textheight-9\baselineskip\relax,keepaspectratio]{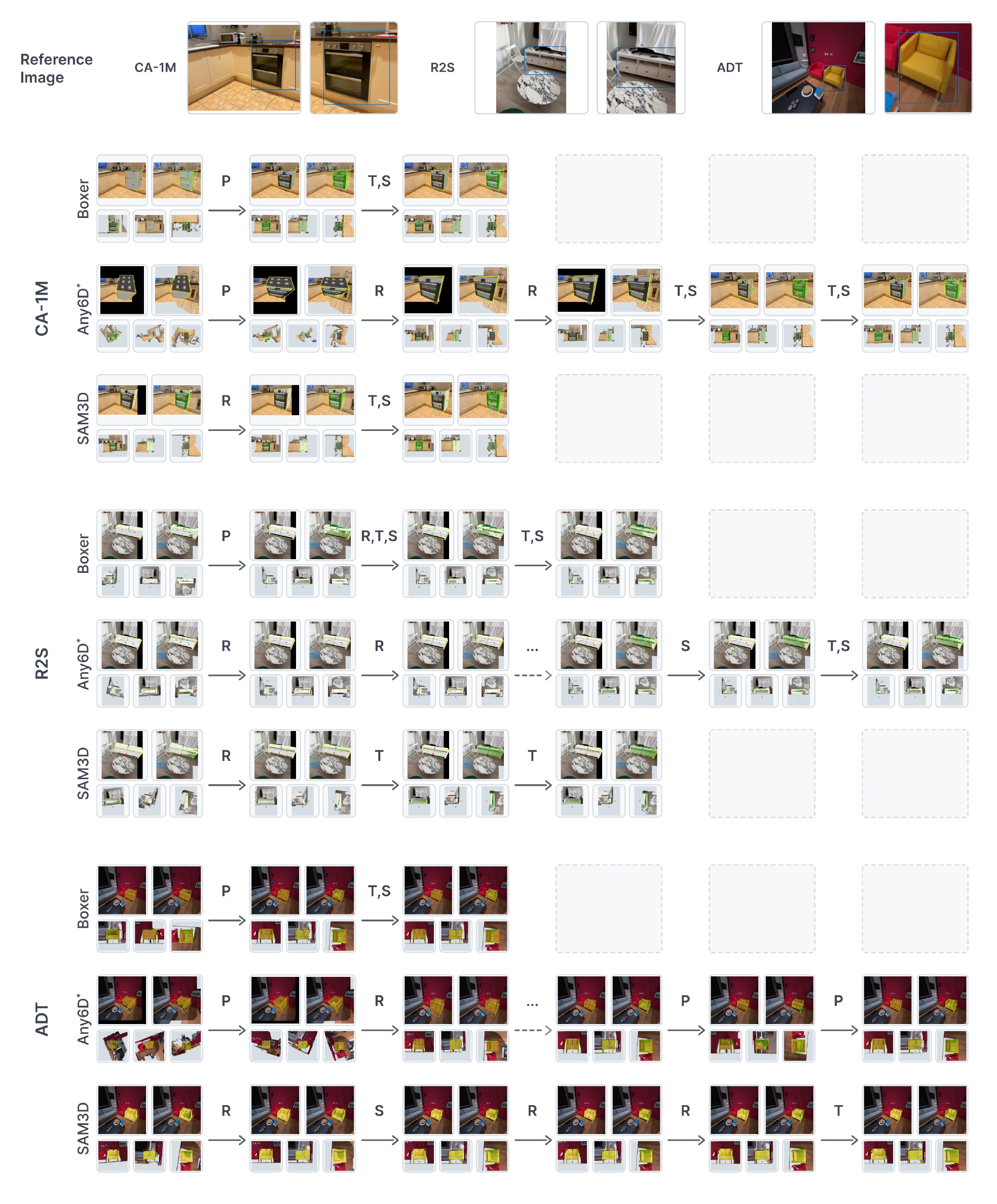}
    \caption{\textbf{GizmoAct pose-refinement trajectories under different initializations.} For each dataset we show a reference image with the target instance boxed, and closed-loop rollouts of the same policy initialized by Boxer, Any6D*, and SAM~3D. Each cell shows the observation at that step (main views on top, orthographic views below); arrow labels denote the predicted action: an axis permutation (P) or the rotation (R), translation (T), and scale (S) components of \texttt{update\_pose}. The final \texttt{stop} action is not shown. Any6D* uses only the depth-and-mask pose initialization from Any6D, without FoundationPose prediction or refinement. Across object categories and geometries, GizmoAct consistently refines each initialization to an accurate pose, outperforming the baseline estimates.}
    \label{fig:qualitative-trajectory-rank-1}
\end{figure}

\paragraph{Qualitative analysis.}
The comparisons in \Cref{fig:qualitative-compare-rank-1-2} cover objects with different categories, shape complexity, and viewpoint difficulty, where GizmoAct aligns the generated 3D models more closely with the reference objects than the baseline methods. We provide more qualitative results in \Cref{fig:qualitative-compare-rank-3-4}. \Cref{fig:qualitative-trajectory-rank-1} further shows that GizmoAct accommodates different pose initializations, which we quantify above; additional trajectory examples are provided in \Cref{fig:qualitative-trajectory-rank-2,fig:qualitative-trajectory-rank-3}.

\begin{figure}[t]
    \centering
    \includegraphics[width=\linewidth]{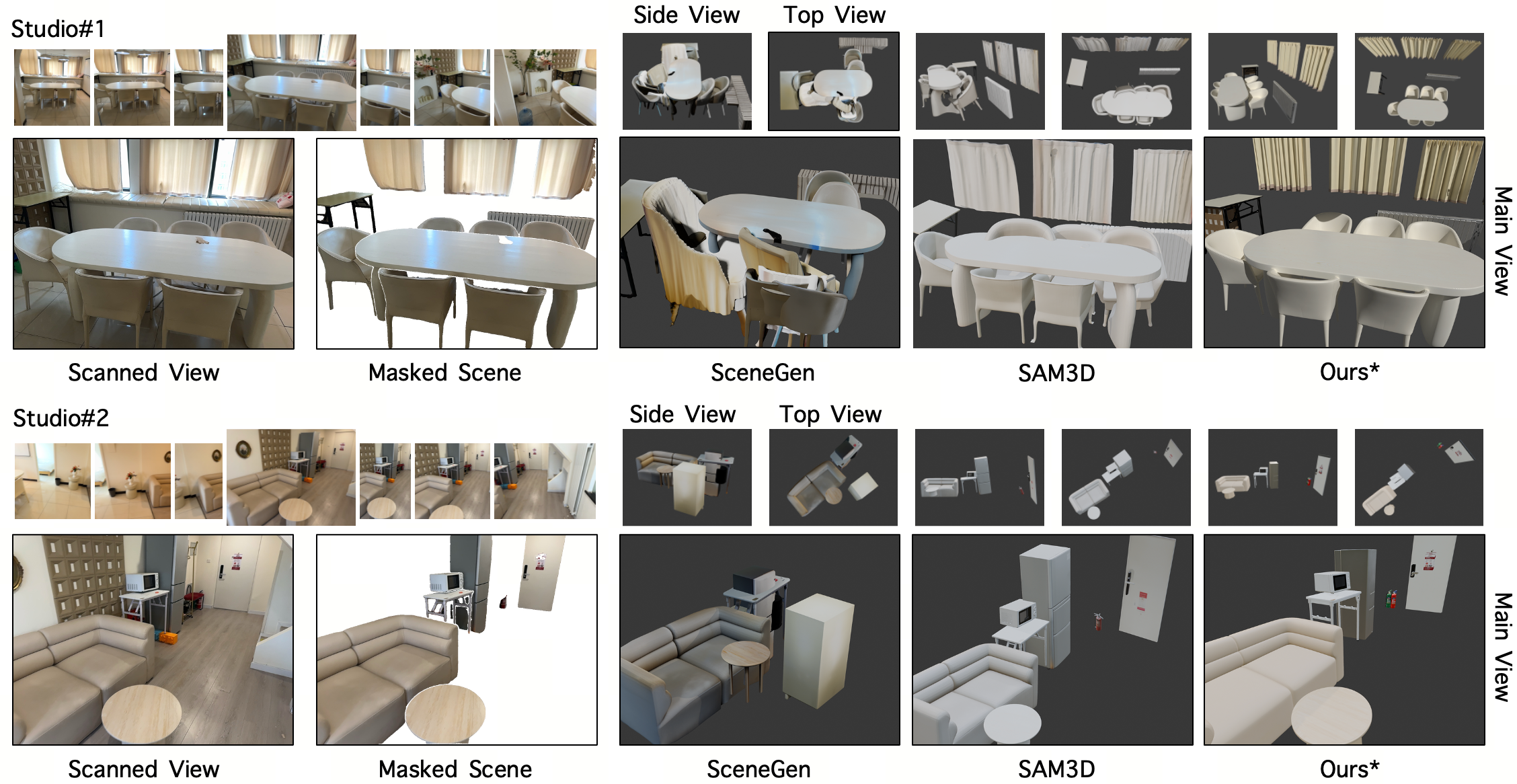}
    \caption{\textbf{Qualitative scene-level comparison.} For each scene, we show the multi-view capture, a scanned reference view, the selected evaluation instances, and reconstructions rendered from the main, side, and top viewpoints. SceneGen and SAM~3D receive the same reference image and instance masks, whereas \ourwork uses per-instance multi-view evidence. The asterisk marks our reconstruction restricted to the object instances included in the comparison.}
    \label{fig:scene-level}
\end{figure}

\subsection{Scene Reconstruction}
\label{sec:exp-scene-reconstruction}

\begin{table}[t]
\caption{\textbf{Scene reconstruction on R2S-Scene.} Scene CD uses squared Euclidean distances in metric scene coordinates and is reported in $\mathrm{m}^2$; total CD is the sum of the two directed terms. Object CD is computed after per-object normalization and is dimensionless. Best in bold, second best underlined.}
\label{tab:scene-object-reconstruction}
\centering
\resizebox{\linewidth}{!}{%
  \begin{tabular}{lccccccc}
    \toprule
    & \multicolumn{3}{c}{Scene CD ($\mathrm{m}^2$) $\downarrow$}
    & & & & \\
    \cmidrule(lr){2-4}
    Method
    & Pred-GT CD
    & GT-Pred CD
    & CD
    & Scene F-Score $\uparrow$
    & BBox IoU $\uparrow$
    & Object CD $\downarrow$
    & Object F-Score $\uparrow$ \\
    \midrule
    SceneGen
    & 0.064
    & 0.364
    & 0.428
    & 0.351
    & 0.092
    & 0.071
    & 0.553 \\

    SAM~3D
    & \underline{0.012}
    & \underline{0.010}
    & \underline{0.022}
    & \underline{0.794}
    & \underline{0.396}
    & \underline{0.038}
    & \underline{0.704} \\

    Ours
    & \textbf{0.006}
    & \textbf{0.004}
    & \textbf{0.010}
    & \textbf{0.924}
    & \textbf{0.495}
    & \textbf{0.034}
    & \textbf{0.736} \\
    \bottomrule
  \end{tabular}%
}
\end{table}

This experiment evaluates the complete system through scene reconstruction.
All methods reconstruct the same selected object instances while independently generating their 3D geometry and recovering their placement from visual observations.
Unlike Sec.~\ref{sec:exp-video-layout}, which evaluates object boxes before asset generation, this experiment jointly measures reconstructed object geometry and final scene layout.

\paragraph{Baselines.}
We compare against SceneGen~\citep{meng2025scenegen} and SAM~3D~\citep{sam3dteam2025sam3d}, which reconstruct compositional object-level scenes from real-image observations.
SceneGen jointly generates object-level assets and predicts their relative transformations in a single feedforward pass, whereas SAM~3D reconstructs prompted instances independently and then assembles them into a scene.
SceneGen and SAM~3D receive the same reference image and instance masks, while our method uses the per-instance multi-view evidence from Sec.~\ref{sec:evidence} for amodal generation and closed-loop grounding.

\paragraph{Datasets and metrics.}
We evaluate on R2S-Scene, the real-world scene split of our R2S benchmark introduced in Sec.~\ref{sec:exp-video-layout}.
We sample 10,240 surface points per object and apply a single rigid FilterReg alignment to the complete SceneGen prediction.
SAM~3D and our method use the same estimated depth maps to place objects in the metric scene coordinate frame.
At the scene level, we report the two directed Chamfer distance (CD) terms; prediction-to-GT CD also penalizes misplaced geometry, while GT-to-prediction CD penalizes missing geometry.
Scene-level CD uses squared Euclidean distances in the metric scene coordinate frame and is reported in $\mathrm{m}^2$; the symmetric CD is the sum of the two directed terms.
We also report the scene F-Score at a distance threshold of 0.1~m and the mean 3D IoU between the axis-aligned bounding boxes of corresponding objects.
At the object level, each predicted--ground-truth pair is normalized and aligned independently before computing CD and F-Score, and the resulting scores are averaged over objects.
Object CD is therefore dimensionless.
The object-level metrics emphasize normalized asset-shape quality, whereas the scene-level metrics additionally reflect object scale, orientation, and relative placement.

\paragraph{Results.}
As shown in \Cref{tab:scene-object-reconstruction}, our method performs best across all reported metrics.
At the scene level, it reduces CD from 0.022 for SAM~3D to 0.010 and increases F-Score from 0.794 to 0.924.
BBox IoU further improves from 0.396 to 0.495, indicating more accurate object scale and relative placement in the composed scene.
At the object level, independently normalized and aligned assets improve from 0.038 to 0.034 in CD and from 0.704 to 0.736 in F-Score.
Both directed scene CD terms are lower than those of SAM~3D, indicating fewer spurious or misplaced surfaces and better ground-truth coverage.

In \Cref{fig:scene-level}, SceneGen exhibits large placement and scale errors, with substantial interpenetration among objects and layouts that deviate from the captured scenes.
SAM~3D better recovers the coarse arrangement visible from the reference image.
However, the side and top views reveal rotation errors that are less apparent from the reference viewpoint.
Our method maintains more consistent object geometry and relative placement across the reference, side, and top views, with fewer visible intersections and closer agreement with the captured scenes.

\FloatBarrier

\subsection{Ablation Studies}
\label{sec:exp-ablation}

\subsubsection{Multi-View Object-Centric Scene Parsing}

\enlargethispage{\baselineskip}

\begin{table}[t]
  \centering
  \caption{\textbf{Ablation of the scene-parsing components on R2S-Scene.} Each variant replaces geometry-aware keyframe selection with temporal uniform sampling or disables object-centric full-sequence evidence consolidation or relation-aware scene refinement. We report scene-level 3D object detection on R2S-Scene and scene reconstruction on an R2S-Scene subset after the same downstream generation and grounding pipeline. Best in bold.}
  \label{tab:ablation-study}
  \resizebox{\linewidth}{!}{%
    \begin{tabular}{lcccccccc}
      \toprule
      & \multicolumn{4}{c}{Scene-Level 3D Object Detection}
      & \multicolumn{4}{c}{Scene Reconstruction} \\
      \cmidrule(lr){2-5}
      \cmidrule(lr){6-9}
      & & & &
      & \multicolumn{3}{c}{Scene CD $\downarrow$}
      & \\
      \cmidrule(lr){6-8}
      Method
      & AP@25 $\uparrow$
      & AP@50 $\uparrow$
      & mAP $\uparrow$
      & F-Score $\uparrow$
      & Pred-GT
      & GT-Pred
      & CD
      & Scene F-Score $\uparrow$ \\
      \midrule

      Full \ourwork
      & \textbf{0.616}
      & \textbf{0.429}
      & \textbf{0.597}
      & \textbf{0.670}
      & \textbf{0.008}
      & 0.011
      & \textbf{0.019}
      & \textbf{0.831} \\

      Uniform keyframe selection
      & 0.547
      & 0.367
      & 0.516
      & 0.635
      & 0.015
      & 0.020
      & 0.035
      & 0.727 \\

      w/o object-centric full-sequence evidence consolidation
      & 0.552
      & 0.394
      & 0.535
      & 0.638
      & 0.011
      & 0.025
      & 0.036
      & 0.734 \\

      w/o relation-aware scene refinement
      & 0.547
      & 0.383
      & 0.526
      & 0.624
      & 0.010
      & \textbf{0.010}
      & 0.020
      & 0.794 \\

      \bottomrule
    \end{tabular}%
  }
\end{table}

\paragraph{Experimental setup.}
We ablate the three stages of Sec.~\ref{sec:evidence}.
Scene reconstruction is evaluated on a subset of R2S-Scene.
First, we replace geometry-aware keyframe selection with temporal uniform sampling.
Second, we disable object-centric full-sequence evidence consolidation, including the propagation of object observations to additional frames and the multi-view validation of their representative 3D estimates.
Third, we disable relation-aware scene refinement, including identity consolidation, missing-support recovery, and support-surface correction.
All variants otherwise share the same downstream generation, grounding, and evaluation protocol.

\paragraph{Results.}
\Cref{tab:ablation-study} reports the effects of the three scene-parsing stages.
Replacing geometry-aware keyframe selection with uniform sampling reduces mAP from $0.597$ to $0.516$ and scene F-Score from $0.831$ to $0.727$, consistent with geometry-aware selection retaining more useful viewpoints.
Disabling object-centric full-sequence evidence consolidation reduces mAP to $0.535$ and scene F-Score to $0.734$.
Its higher GT-to-prediction CD ($0.025$ versus $0.011$) is consistent with reduced scene coverage when observations beyond the selected keyframes are not consolidated.

Without relation-aware scene refinement, mAP decreases to $0.526$ and scene F-Score to $0.794$, while total scene CD changes only slightly.
Relation-aware scene refinement therefore improves the recovered 3D boxes and scene F-Score, while having little effect on total scene CD.
Overall, all three ablations reduce both detection mAP and scene F-Score, with uniform keyframe sampling producing the largest decrease.

\subsubsection{GizmoAct Training}

\begin{table}[t]
    \caption{\textbf{Effect of the RL training pose distribution.} The two rows differ only in the initial-pose distribution used during RL training: poses produced by Boxer versus random perturbations. Both policies are evaluated from Boxer initialization, and matching the training distribution to the test-time initializer improves every reported metric. @0.05 denotes the ADD-SB success rate at that threshold; best per dataset in bold.}
    \label{tab:gizmoact-rl-pose-distribution}
    \centering
    \resizebox{0.92\linewidth}{!}{%
    \begin{tabular}{lccccccccc}
        \toprule
        & \multicolumn{3}{c}{R2S-Object}
        & \multicolumn{3}{c}{CA-1M}
        & \multicolumn{3}{c}{ADT} \\
        \cmidrule(lr){2-4}\cmidrule(lr){5-7}\cmidrule(lr){8-10}
        Training poses & ADD-SB $\downarrow$ & @0.05 $\uparrow$ & 3D IoU $\uparrow$
        & ADD-SB $\downarrow$ & @0.05 $\uparrow$ & 3D IoU $\uparrow$
        & ADD-SB $\downarrow$ & @0.05 $\uparrow$ & 3D IoU $\uparrow$ \\
        \midrule
        Boxer initialization & \textbf{0.017} & \textbf{92.0\%} & \textbf{0.719} & \textbf{0.021} & \textbf{83.4\%} & \textbf{0.607} & \textbf{0.021} & \textbf{90.0\%} & \textbf{0.670} \\
        Random perturbation & 0.022 & 90.4\% & 0.678 & 0.026 & 77.9\% & 0.568 & 0.025 & 81.7\% & 0.626 \\
        \bottomrule
    \end{tabular}}
\end{table}

\begin{figure}[t!]
    \centering
    \includegraphics[width=\linewidth]{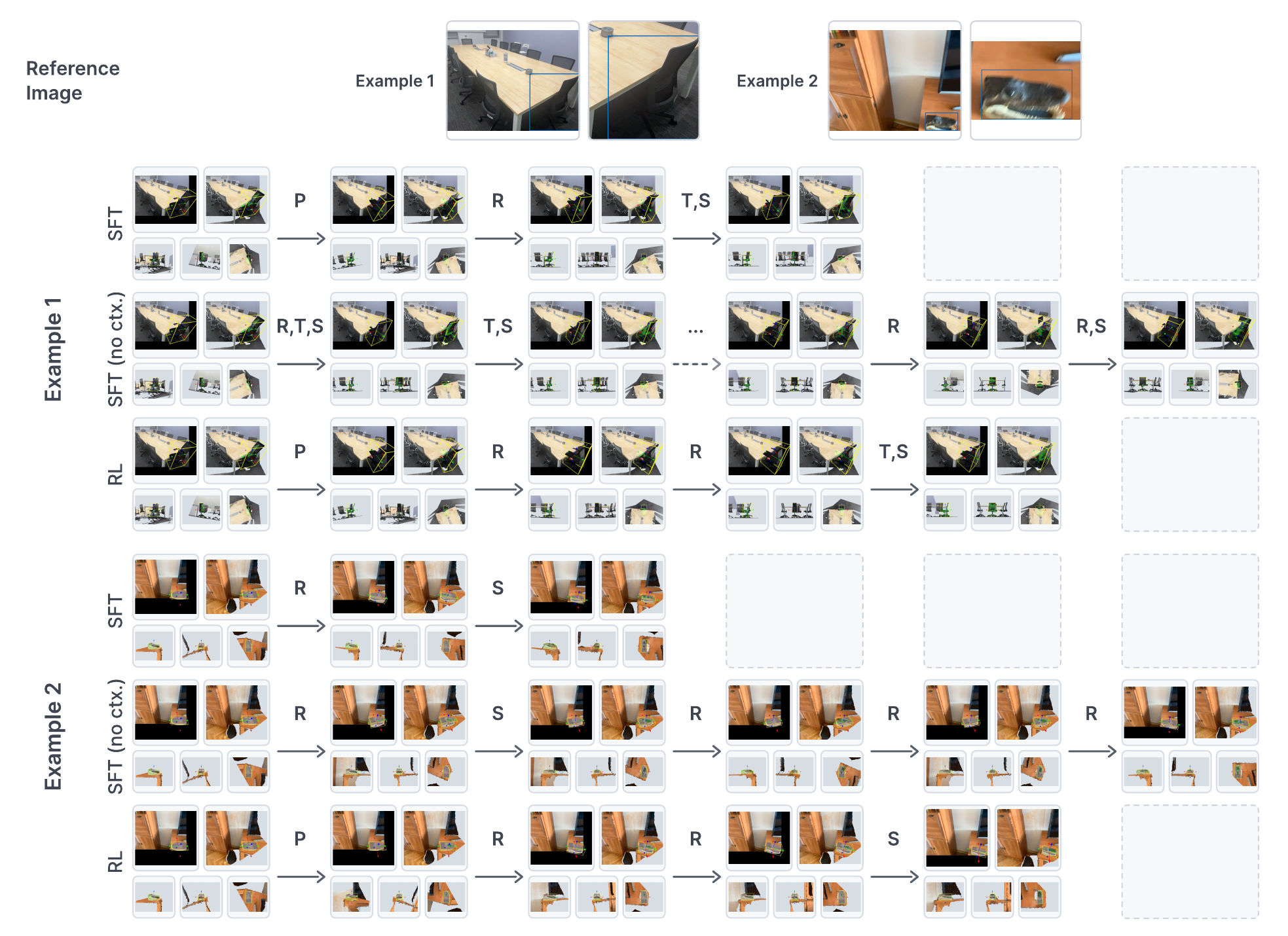}
    \caption{\textbf{Qualitative comparison of GizmoAct training strategies on two difficult examples.} Each row shows the closed-loop rollout of the SFT, SFT (no context), or RL policy from the same initialization, evaluated as in \Cref{tab:gizmoact-training-strategy}; notation follows \Cref{fig:qualitative-trajectory-rank-1}. The SFT policies stop early with residual rotation errors or drift without converging, whereas RL keeps correcting the remaining rotation before stopping.}
    \label{fig:gizmoact-training-comparison}
    \vspace{0.8\baselineskip}
    \captionof{table}{\textbf{Comparison of GizmoAct training strategies on the hard subset.} SFT (no context) is trained with single-step supervision and no interaction history; all three policies are evaluated in the same closed-loop setting with at most 12 steps. RL improves every metric, with the largest gains on rotation accuracy. Evaluated on a hard subset of 110 asymmetric objects with large Any6D* initialization errors; best in bold.}
\label{tab:gizmoact-training-strategy}
\centering
\resizebox{0.92\linewidth}{!}{%
\begin{tabular}{lcccccc}
    \toprule
    Method & ADD-SB $\downarrow$ & @0.05 $\uparrow$ & @0.01 $\uparrow$ & 3D IoU $\uparrow$ & Rot. error ($^\circ$) $\downarrow$ & Rot.@$5^\circ$ $\uparrow$ \\
    \midrule
    GizmoAct (SFT) & 0.040 & 70.0\% & 6.4\% & 0.612 & 45.19 & 52.7\% \\
    GizmoAct (SFT, no context) & 0.035 & 86.4\% & 0.9\% & 0.636 & 27.62 & 60.9\% \\
    GizmoAct (RL) & \textbf{0.032} & \textbf{87.3\%} & \textbf{11.8\%} & \textbf{0.685} & \textbf{20.98} & \textbf{67.3\%} \\
    \bottomrule
\end{tabular}}

\end{figure}

\paragraph{Pose distribution for RL training.}
During RL, we can specialize the policy to the pose errors produced by the initializer used at inference. We train one policy from Boxer-initialized poses and another from random pose perturbations, then evaluate both from Boxer initialization on R2S-Object, CA-1M, and ADT. As shown in \Cref{tab:gizmoact-rl-pose-distribution}, matching the training distribution to Boxer initialization improves every reported metrics: ADD-SB@0.05 increases by 1.6, 5.5, and 8.3 percentage points on R2S-Object, CA-1M, and ADT, respectively, and 3D IoU improves on all three datasets. These gains show that we can steer the policy toward the error distribution of a target initializer by training on poses produced by that initializer.

\paragraph{RL versus SFT.}
The gains from RL over SFT are most evident on difficult cases, where the policy must recover from large initialization errors. We therefore construct a hard subset of 110 asymmetric objects from R2S-Object, CA-1M, and ADT, selecting samples with large errors from the Any6D-style depth-and-mask initializer; all policies are evaluated on these large-error initial poses. Beyond the metrics of Sec.~\ref{sec:exp-object-pose}, we report the stricter ADD-SB@0.01, together with the geodesic rotation error (Rot. error) and the fraction of samples below $5^\circ$ (Rot.@$5^\circ$); the rotation metrics are unambiguous here because the subset excludes symmetric objects. We compare the RL policy against two SFT baselines. The first is the SFT policy described in Sec.~\ref{sec:gizmoact}, which is also used to initialize RL; we denote it GizmoAct (SFT). The second is trained separately without interaction history: given only the current observation, it predicts either a complete pose correction or \texttt{stop}. We denote this variant GizmoAct (SFT, no context). Although the second baseline is trained with single-step supervision, all three policies are evaluated in the same closed-loop setting with at most four views and 12 refinement steps. \Cref{tab:gizmoact-training-strategy} shows that GizmoAct (RL) lowers rotation error to $20.98^\circ$ from $45.19^\circ$ and $27.62^\circ$ for the two SFT baselines, while raising Rot.@$5^\circ$ to 67.3\% from 52.7\% and 60.9\%. \Cref{fig:gizmoact-training-comparison} shows the same pattern qualitatively: RL continues to correct residual rotation errors that persist in the SFT trajectories, reducing cases in which position and scale are correct but rotation is not.

\section{Related Work}

\subsection{Composable 3D Scene Modeling}

Composable scene modeling recovers a captured scene as separable, editable object assets rather than a monolithic reconstruction.
Optimization-based methods attach object structure to a differentiable-rendering reconstruction \citep{schonberger2016sfm, mildenhall2020nerf, kerbl20233d} through object discovery \citep{yu2021uorf}, mask supervision \citep{yang2021objectnerf, wu2022objectsdf, wu2023objectsdfplusplus}, label lifting \citep{zhi2021semanticnerf, kundu2022pnf, siddiqui2022panopticlifting}, or Gaussian grouping \citep{ye2024gaussiangrouping}, but explain only the observed pixels, leaving occluded surfaces unrecovered.
End-to-end methods map an image to a decomposed scene, from holistic layout--pose--shape regression \citep{huang2018cooperative, nie2020total3d, zhang2021holistic} to multi-instance and part-level latent diffusion \citep{yan2024frankenstein, huang2024midi, meng2025scenegen, sam3dteam2025sam3d, lin2025partcrafter, chen2025autopartgen}; synthetic scene supervision limits generalization to real captures, and joint shape--pose prediction couples both errors \citep{sam3dteam2025sam3d, zadaianchuk2026recgen}.

Closest to our setting, multi-stage pipelines decompose the task into segmentation, per-object asset acquisition, and placement.
Early pipelines retrieve and align CAD models \citep{avetisyan2018scan2cad, avetisyan2020scenecad, kuo2020mask2cad, gumeli2021roca, gao2023diffcad, dai2024acdc, torne2024rialto}, capping fidelity at database coverage; recent ones generate each object with image-to-3D models \citep{ardelean2024gen3dsr, zhou2024deeppriorassembly}, anchored to sensor depth \citep{agarwal2024scenecomplete}, completed amodally in 3D \citep{wu2025amodal3r, dong2025hiscene}, constrained by inter-object relations and physics \citep{yao2025cast, xia2025holoscene}, and scaled to scan datasets or video \citep{yu2025metascenes, xia2026simrecon}.
Yet the stages run in a fixed order, each presuming precise inputs---clean masks, unoccluded crops, depth that agrees with the generated asset---so an early error is inherited by every later stage with no mechanism to correct it.
\ourwork keeps the modular decomposition but relocates precision to the output end: parsing gathers per-instance multi-view evidence, generation conditions on that evidence, and GizmoAct closes the loop on placement, absorbing mismatched assets and coarse initializations---so upstream stages need only be approximately right on cluttered real captures.

\subsection{Scene-Level 3D Object Detection}

Scene-level 3D object detection localizes every object instance of a captured scene as an oriented 3D box in a single de-duplicated inventory, the output of the parse step in \ourwork.
Monocular detectors predict boxes from a single frame at growing scale and vocabulary \citep{brazil2023omni3d, yao2024ovmono3d, zhang2025detany3d, lazarow2024cubify, wilddet3d2026}, but remain fragile to occlusion and viewpoint and give no instance persistence across views.
Multi-view methods either pool posed views into a scene volume \citep{rukhovich2022imvoxelnet, xu2023nerfdet} or---closest to our setting---detect per view and fuse boxes afterwards: Boxer \citep{boxer2026} merges per-view lifts with rule-based geometric fusion, BoxFusion \citep{lan2025boxfusion} fuses frozen detections in real time, and Rooms from Motion \citep{lazarow2025roomsfrommotion} trains an object matcher that jointly recovers tracks and camera poses.

A complementary line associates 2D observations into object-level maps, spanning object-centric SLAM \citep{mccormac2018fusionpp, kong2023vmap}, multi-view lifting of foundation-model masks \citep{yang2023sam3d, takmaz2023openmask3d, yan2024maskclustering, xu2024embodiedsam}, and open-vocabulary scene graphs \citep{gu2024conceptgraphs, werby2024hovsg}.
Across both lines, instances are associated by geometric overlap and feature similarity, which cluttered captures undermine: repeated near-identical instances risk being merged or duplicated, occluded objects missed, and the inventory rarely re-examined.
The parse step of \ourwork instead detects on keyframes, fuses each instance's multi-view observations, and verifies and completes the scene graph against the full capture---removing spurious instances, correcting implausible support relations, and disambiguating near-identical objects---yielding the highest detection mAP in all reported settings; each verified node moreover carries the multi-view evidence bundle that asset generation and placement consume.

\subsection{3D Visual Grounding and Iterative Pose Refinement}

3D visual grounding localizes a referred object, classically via closed-vocabulary specialist detectors over reconstructed point clouds \citep{chen2019scanrefer, achlioptas2020referit3d, zhao2021threeDVG, jain2021butddetr, wu2023eda, zhang2023multi3drefer}.
Foundation models change the interface, yet LLMs with 3D-native inputs generalize weakly beyond scene-scan corpora \citep{hong20233dllm, zhu20233dvista, zhu2024llava3d}, 2D-input VLMs trained with spatial and 3D-grounding supervision localize imprecisely \citep{chen2024spatialvlm, cho2024cubellm, bai2025qwen3vl, yang2025vst, fang2025robix}, and agentic systems that orchestrate grounding tools stop once the referent is identified, leaving its coarse box unrefined \citep{yang2023llmgrounder, xu2024vlmgrounder, li2024seeground}.
Most related to \ourwork, VLM agents place and arrange assets in 3D scenes: PlaceIt3D \citep{abdelreheem2025placeit3d} benchmarks the task, FirePlace \citep{huang2025fireplace} solves single-shot common-sense constraints, VULCAN \citep{kuang2026vulcan} iterates through code-level tools, and VLMPose \citep{baik2026vlmpose} closes a 6-DoF loop against re-renderings.
These methods optimize plausibility under language constraints, and none emphasizes precise alignment to the observed point cloud.
GizmoAct addresses the opposite regime---grounding a generated asset to one observed instance with metric 9-DoF alignment to the point cloud---and its GUI abstraction generalizes beyond object pose: the same loop manipulates any 3D proxy (a 3D box for detection and layout, a gizmo frame for orientation estimation \citep{wang2024orientanything}) and consumes multi-view and virtual renderings for robustness.

Aligning a 3D model to observations is the classic object pose estimation problem, typically solved by pose initialization followed by iterative refinement.
Render-and-compare methods regress a relative pose update from the rendered--observed pair \citep{li2018deepim, labbe2020cosypose, cifka2024focalposepp}, scaled to novel objects by MegaPose \citep{labbe2022megapose} and FoundationPose \citep{wen2023foundationpose}; differentiable-rendering variants descend photometric residuals \citep{lin2020inerf, tremblay2023diffdope}, and Any6D \citep{lee2025any6d} admits generated meshes.
Their updates are inferred by comparing the rendering with the observation, presuming the rendered model can resemble the observed object; convergence is decided externally by a fixed iteration count \citep{li2018deepim, wen2023foundationpose}, and scale beyond SE(3) is rarely estimated.
A second family casts refinement as sequential decision making with small reinforcement-learned policies \citep{krull2017poseagent, shao2021pfrl}: discrete incremental moves with a learned stop action \citep{busam2020ilikemoveit}, and imitation-then-RL refinement over discrete registration steps \citep{bauer2021reagent, bauer2022sporeagent, rohrl2023trackagent, liu2023artperl}.
GizmoAct inherits this decision structure---incremental actions, learned stopping, imitation followed by reinforcement learning---but upgrades the policy to a pretrained VLM operating a GUI: it manipulates the object through its gizmo, decides its own termination from the rendering alone, tolerates geometry that disagrees with the observation, and refines the full 9-DoF state including anisotropic scale.

\section{Conclusion}

This paper introduces \ourwork, a composable scene modeling system that keeps the parse--generate--place order but redistributes the requirements along it, so each step consumes only what a real capture reliably provides. The parse step discovers candidate instances on geometry-aware keyframes, consolidates their evidence over the full sequence, and reasons over inter-object relations to refine the resulting scene graph. The generate step completes each instance amodally from its evidence, and GizmoAct, a VLM policy that casts placement as multi-turn GUI interaction, aligns each generated asset by closed-loop gizmo manipulation and decides itself when alignment is reached. Precision is reached at the closed-loop end of the pipeline, so the upstream steps tolerate the occlusion, clutter, and asset--observation mismatch that real indoor captures impose.

The experiments evaluate \ourwork at three levels. For scene-level 3D object detection, the scene graph surpasses prompt-based 3D detectors on CA-1M and R2S-Scene. For object pose estimation, GizmoAct improves strict alignment and 3D IoU on R2S-Object, CA-1M, and ADT, and a single policy handles initializers with different error profiles without retraining. At the system level, the composed scenes improve scene F-Score from 0.794 for SAM~3D to 0.924. A remaining limitation is that objects that remain missing after scene parsing cannot be recovered by the subsequent generation or grounding stages. Moreover, agentic refinement currently operates only at the final placement step; extending this closed-loop treatment across the whole parse--generate--place pipeline, toward a fully agentic framework, is an important direction for future work.

\phantomsection
\section*{Contributions}
\label{sec:contributions}

\noindent\textbf{Authors} Minghan Qin$^{1,\ddagger}$, Yuang Wang$^{1,\ddagger}$, Xiuyu Yang$^{1,*,\ddagger}$, Yushi Long$^{1,3,*}$, Yujian Zhang$^{1,3,*}$, Ruihuan Wang$^{1,2,*}$, Kai Ye$^{1,2,*}$, Yangang Zhang$^{1,\dagger}$, Hang Li$^{1}$

\noindent\textbf{Affiliations} $^{1}$ByteDance Seed, $^{2}$Peking University, $^{3}$Zhejiang University

\noindent\textbf{Acknowledgments} We thank Wei Li, Heng Dong, and Qifeng Zhang for providing the pretrained VLM and for their help with model training. We thank Lihao Liu, Baifeng Xie, and Yiming Qiao for their help with simulation data, evaluation, and physical-plausibility post-processing.

\noindent$^{\ddagger}$ Equal contribution\\
$^{*}$ Work done during internship at ByteDance Seed\\
$^{\dagger}$ Corresponding author

\clearpage
\begin{minipage}{\textwidth}
    \centering
    \includegraphics[width=\linewidth,height=\dimexpr\textheight-6\baselineskip\relax,keepaspectratio]{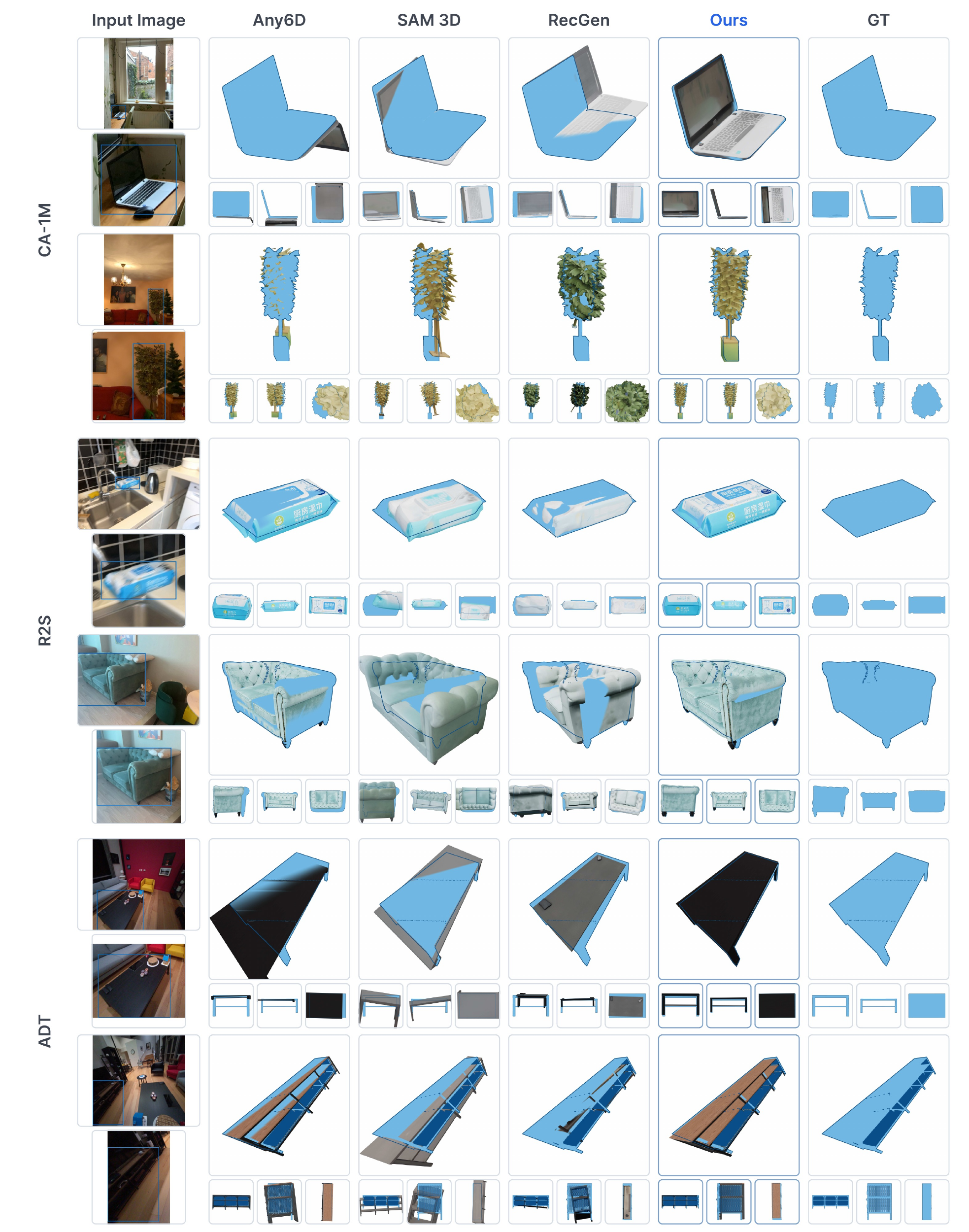}
    \captionof{figure}{\textbf{Additional qualitative comparison of object pose estimation on CA-1M, R2S-Object, and ADT.} Same layout and reading protocol as \Cref{fig:qualitative-compare-rank-1-2}: each method's posed model is rendered jointly with the ground-truth posed model (blue), from a main view and three axis-aligned close-ups.}
    \label{fig:qualitative-compare-rank-3-4}
\end{minipage}
\clearpage

\begin{minipage}{\textwidth}
    \centering
    \includegraphics[width=\linewidth]{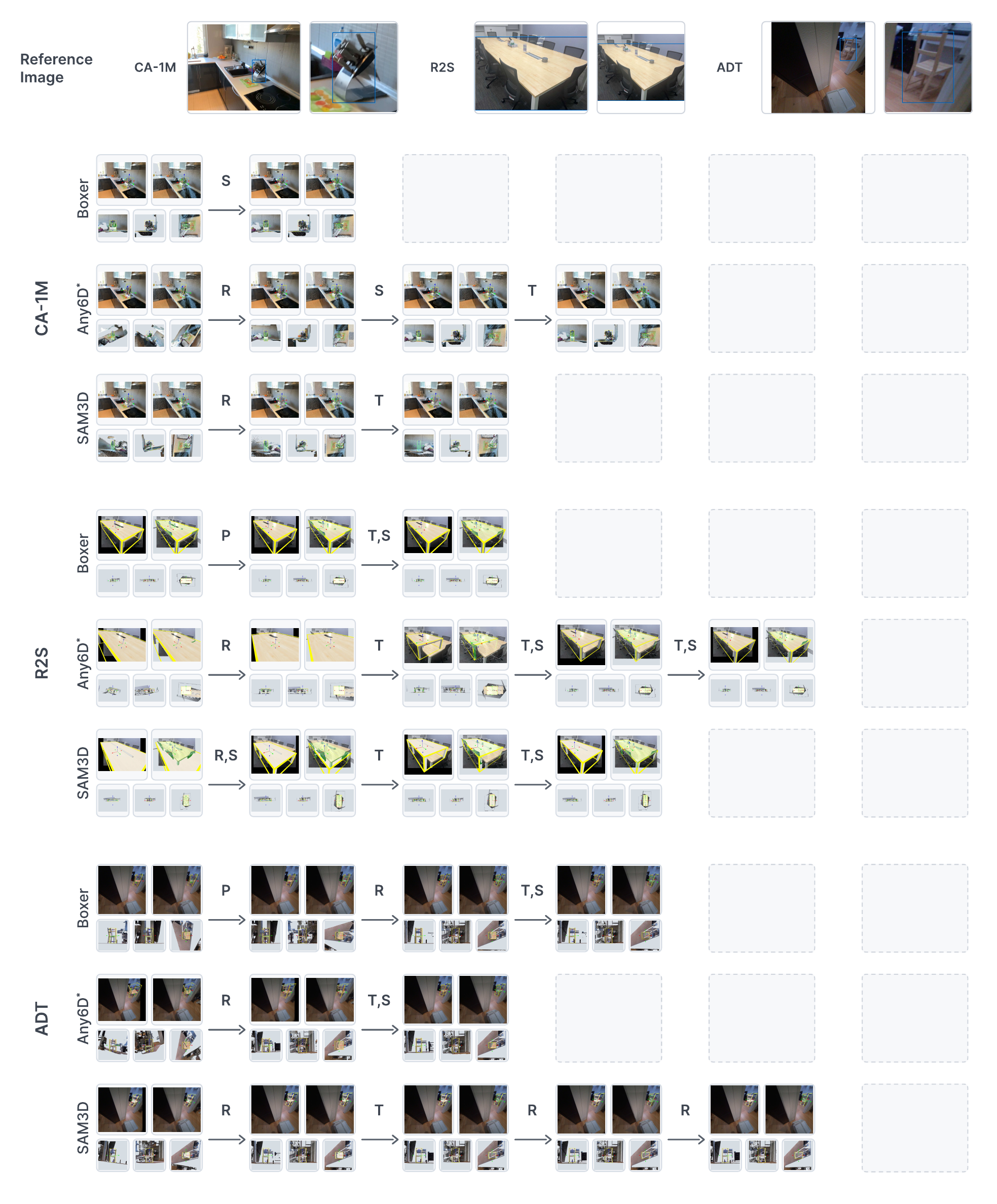}
    \captionof{figure}{\textbf{Additional GizmoAct trajectory examples on CA-1M, R2S-Object, and ADT.} Same layout and notation as \Cref{fig:qualitative-trajectory-rank-1}: a reference image per dataset, followed by closed-loop rollouts from Boxer, Any6D*, and SAM~3D initializations.}
    \label{fig:qualitative-trajectory-rank-2}
\end{minipage}
\clearpage

\begin{minipage}{\textwidth}
    \centering
    \includegraphics[width=\linewidth]{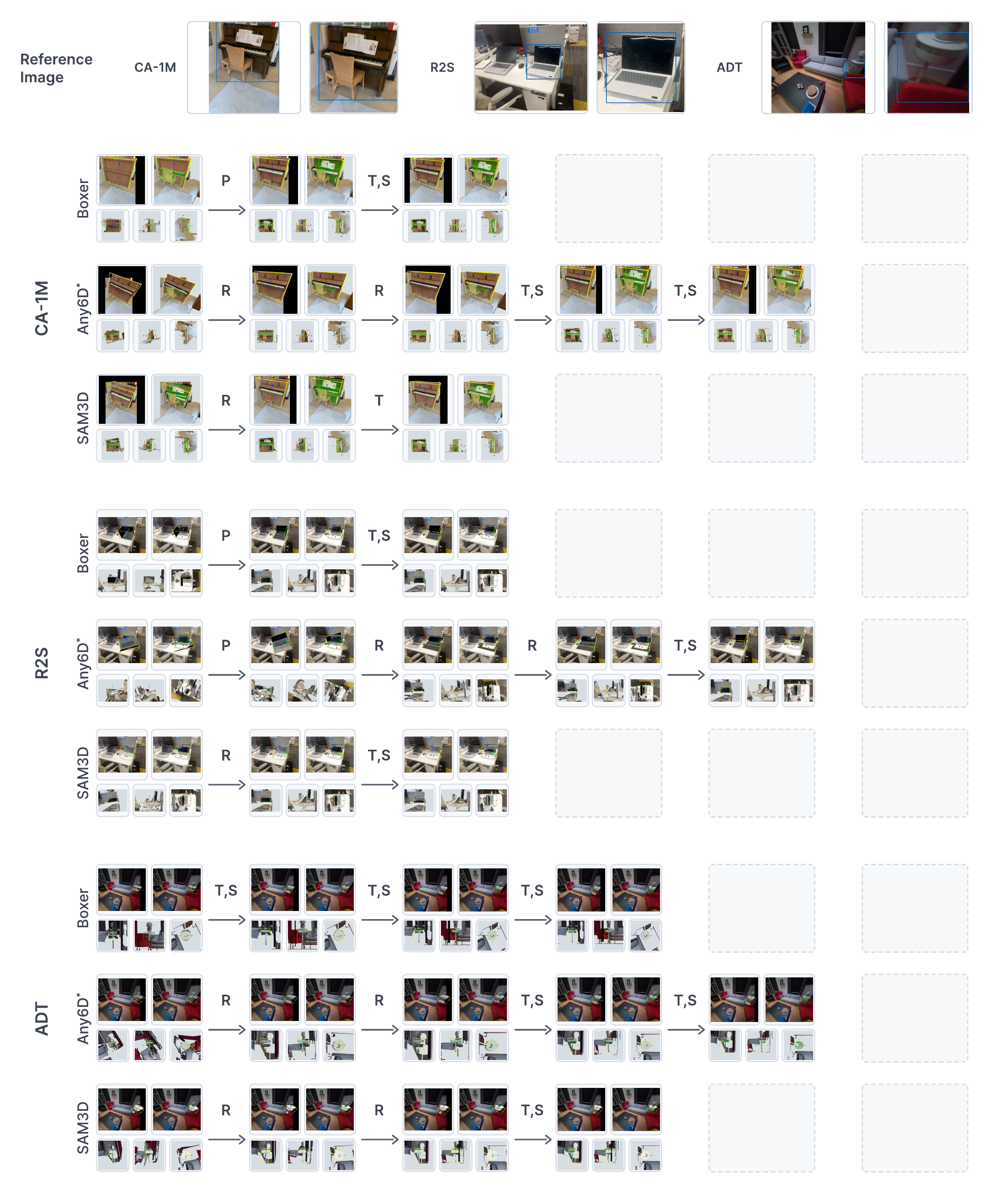}
    \captionof{figure}{\textbf{Further GizmoAct trajectory examples on CA-1M, R2S-Object, and ADT.} Same layout and notation as \Cref{fig:qualitative-trajectory-rank-1}: a reference image per dataset, followed by closed-loop rollouts from Boxer, Any6D*, and SAM~3D initializations.}
    \label{fig:qualitative-trajectory-rank-3}
\end{minipage}
\clearpage

\clearpage
\bibliographystyle{plainnat}
\bibliography{main}

\end{document}